%% file: hypervl.tex
\documentclass{article}

\usepackage[final]{neurips_2024}

\usepackage[utf8]{inputenc} % allow utf-8 input
\usepackage[T1]{fontenc}    % use 8-bit T1 fonts
\usepackage{hyperref}       % hyperlinks
\usepackage{url}            % simple URL typesetting
\usepackage{booktabs}       % professional-quality tables
\usepackage{amsfonts}       % blackboard math symbols
\usepackage{nicefrac}       % compact symbols for 1/2, etc.
\usepackage{microtype}      % microtypography
\usepackage{hyperref}

\usepackage[table,x11names]{xcolor}
\usepackage{makecell}
\usepackage{graphicx}
\usepackage{multirow}
\usepackage{amsmath}
\usepackage{amssymb} % For \checkmark and other symbols if needed
\usepackage{geometry} % 设置页面边距
\usepackage{tabularx}   % 提供 tabularx 环境和 X 列类型
\usepackage{fixltx2e}
\usepackage{subcaption}
\usepackage{appendix}
\usepackage{array}
\usepackage{ragged2e}
\usepackage{enumitem}
\usepackage{caption}
\usepackage{CJKutf8}
\usepackage{fontawesome5}

\title{HyperVL: An Efficient and Dynamic Multimodal Large Language Model for Edge Devices}

\author{
HyperAI Team, Xiaomi Corporation 
}

\begin{document}
\maketitle

\input{sec/0_abstract}    
\input{sec/1_intro}
\input{sec/2_related}
\input{sec/3_method}

\input{sec/4_experiment}
\input{sec/5_conclusion}
\bibliographystyle{unsrtnat}
\bibliography{hypervl}

\appendix
\input{sec/7_authors}
\input{sec/appendixes}

\end{document}

%% file: sec/0_abstract.tex
\begin{abstract}
% Current multimodal large models rely heavily on cloud-based inference. 
Current multimodal large lanauge models possess strong perceptual and reasoning capabilities, however high computational and memory requirements make them difficult to deploy directly on on-device environments. 
% As the densing law rises, small-parameter models are gradually acquiring general capabilities, making on-device models a growing practical demand. Typical ViT encoders still face bottlenecks such as high latency and large memory consumption when processing high-resolution inputs. 
While small-parameter models are progressively endowed with strong general capabilities, standard Vision Transformer (ViT) encoders remain a critical bottleneck, suffering from excessive latency and memory consumption when processing high-resolution inputs.
To address these challenges, we introduce HyperVL, an efficient multimodal large language model tailored for on-device inference. HyperVL adopts an image-tiling strategy to cap peak memory usage and incorporates two novel techniques: (1) a Visual Resolution Compressor (VRC) that adaptively predicts optimal encoding resolutions to eliminate redundant computation, and (2) Dual Consistency Learning (DCL), which aligns multi-scale ViT encoders within a unified framework, enabling dynamic switching between visual branches under a shared LLM. Extensive experiments demonstrate that HyperVL achieves state-of-the-art performance among models of comparable size across multiple benchmarks.  Furthermore, it significantly significantly reduces latency and power consumption on real mobile devices, demonstrating its practicality for on-device multimodal inference.
\end{abstract}

%% file: sec/1_intro.tex
\section{Introduction}
In recent years, multimodal large language models (MLLMs)—such as GPT-4o, Gemini \citep{gemini,gemini_1.5,gemini_2,gemini_2.5,gemini2p5pro_2025,gemini_3,gemini_3pro}, Claude \citep{claude3_2024,claude4_2025,claude_opus4_1_2025,claude_sonnet4_5_2025}, QwenVL \citep{qwen-vl,qwen2-vl,qwen3-vl}, and InternVL \citep{internvl3,internvl3.5}—have achieved rapid progress, delivering impressive breakthroughs in cross-modal understanding, visual reasoning, optical character recognition (OCR), and open-world perception. At the same time, the demand for multimodal capabilities on on-device platforms is growing rapidly. However, these models are primarily designed for cloud-based inference; their complex architectures and extremely large parameter scales make it challenging to run them efficiently on on-device environments, such as smartphones and tablets, where compute and memory budgets are highly constrained.  On-device models provide strong privacy protection for user data and significantly reduce cloud inference costs, making the performance–efficiency trade-off of on-device MLLMs a central concern for major device manufacturers.

As data quality improves and model architecture design matures, the densing law of LLMs \citep{minicpmv-1-2-2.5} continues to rise, enabling small-parameter models to achieve strong general capabilities. This trend provides a practical foundation for on-device deployment of large models. 
% Nevertheless, existing on-device MLLMs still face considerable challenges. 
Recently, compact MLLMs such as Qwen3 VL 2B \citep{qwen3-vl} and InternVL 2B \citep{internvl3,internvl3.5} have demonstrated competitive performance. Nevertheless, a critical bottleneck remains: these models typically rely on standard Vision Transformer (ViT) encoders, which suffer from quadratic complexity. When processing high-resolution inputs, which is essential in typical on-device scenarios such as user interface (UI) understanding and grounding, they are prone to high memory consumption and noticeable inference latency during on-device deployment. Attempts to mitigate this, such as Apple’s FastVLM \citep{apple-fastvlm}, often resort to reducing visual tokens via aggressive convolutional downsampling, but this frequently comes at the cost of substantial performance degradation.

% We observe that general visual encoders—particularly ViTs under high-resolution inputs—are the primary performance bottleneck in on-device deployment, directly causing elevated latency and memory usage. 
% For example, QwenVL 2B \citep{qwen25-vl} and InternVL 2B \citep{internvl3.5} already deliver competitive performance, yet their reliance on standard Vision Transformer (ViT) encoders makes them prone to high memory consumption and noticeable inference latency during on-device deployment—issues that become particularly severe under high-resolution inputs which is essential in typical on-device scenarios such as user interface (UI) understanding and grounding. Apple’s FastVLM \citep{apple-fastvlm} approach improves efficiency by reducing the number of visual tokens with convolutional networks, but this comes at the cost of substantial performance degradation. %Honor’s recently released MagicVL achieves a better balance between accuracy and efficiency, primarily by shrinking the ViT itself, but it still does not fully explore the optimization potential of on-device inference. 

% (OPPO? vivo?)
\begin{figure*}[!t]
\vskip -0.05in
\centering
\includegraphics[width=0.99\linewidth]{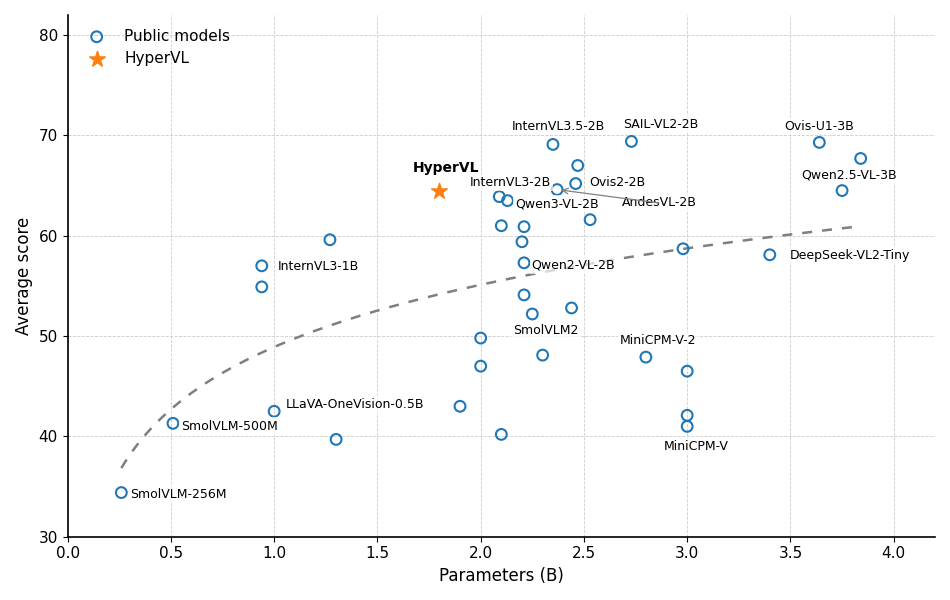}
\caption{OpenCompass Average Score vs. Model Parameters. We compare HyperVL with open‑source models released after October 2024, based on OpenCompass benchmarks. HyperVL (marked by the star) achieves superior performance comparable to larger models while maintaining a compact parameter footprint.}%The dashed line represents the general scaling trend of existing models.}
\label{img:trend}
\vskip -0.1 in
\end{figure*}

To this end, we introduce HyperVL, an efficient multimodal large language model tailored for on-device inference. Specifically, we adopt an image-tiling strategy that divides high-resolution inputs into smaller patches that can be encoded independently, thereby reducing peak memory consumption. Building on this foundation, we further reduce on-device inference costs through two key techniques. First, the Visual Resolution Compressor (VRC) adaptively predicts the optimal visual encoding resolution based on the image information density, avoiding unnecessary high-resolution computation and reducing visual encoding latency. Second, we apply Dual Consistency Learning (DCL) on ViT encoders of different capacities so that they can seamlessly connect to a shared LLM, enabling dynamic switching between lightweight and high-accuracy visual paths depending on task types, latency budget, or device compute capability. 

% In addition, we construct a large-scale, high-information-density screen image dataset and optimize the model for typical on-device tasks such as GUI understanding, app recommendation, and agent action planning, allowing the model to retain broad generality while exhibiting stronger screen reasoning capabilities.

In summary, our contributions are as follows:
\begin{itemize}

\item We propose HyperVL, a multimodal large language model designed for efficient on-device inference. As shown in Fig. \ref{img:trend}, under comparable model sizes, HyperVL achieves state-of-the-art results across multiple authoritative multimodal benchmarks and outperforms existing general models on on-device scenarios, marking a meaningful capability breakthrough for real-world on-device applications.

\item We significantly improve visual encoding efficiency through dual consistency learning for ViTs and the Visual Resolution Compressor. The Compressor is plug-and-play and can be directly integrated into any pretrained MLLM, providing a general acceleration method for on-device multimodal inference.

\item We systematically validate on-device inference performance on real hardware. We quantify the reduction in latency and memory consumption on commercial mobile devices, providing a comprehensive system-level demonstration of an efficient multimodal inference pipeline optimized for on-device deployment.
 
\end{itemize}

%% file: sec/2_related.tex
\section{Related Work}

\textbf{Small-scale Multimodal Models.}
In recent years, multimodal large language models (MLLMs) have achieved remarkable progress across understanding and generation tasks. However, much of this progress has been driven by scaling laws \citep{scaling} where increasing parameter counts—from tens of billions to over a trillion—continues to push performance ceilings but also leads to prohibitive training and deployment costs, making such models challenging to run on-device. Consequently, both academia and industry have begun exploring smaller, more energy-efficient multimodal models. Representative work includes the Qwen-VL \citep{qwen-vl,qwen25-vl,qwen3-vl} and InternVL families \citep{internvl,internvl2,internvl3,internvl3.5}, which provide variants ranging from 1B to 4B parameters and significantly reduce computational complexity while maintaining strong multimodal reasoning capabilities. Nonetheless, their ViT backbones still exhibit high memory consumption and long inference latency during on-device execution, particularly under high-resolution visual inputs. FastVLM \citep{apple-fastvlm} improves ViT inference efficiency by compressing visual tokens using CNNs, though this design introduces noticeable performance degradation. 
%The MiniCPM-V series \citep{minicpmv-1-2-2.5, minicpmv-4.5} performs systematic optimization across architecture, training, and deployment, using a 3D Resampler to reduce redundant visual computation and achieving on-device latency under 200 ms. 
Several manufacturers have successfully deployed small-scale models on real devices via joint hardware–software optimization. AndesVL \citep{oppo-andesvl} proposes a system-level optimization pipeline and utilizes ARM-level kernel inference optimization to enable stable execution of a 3B model within 2 GB of memory. Although these works demonstrate substantial progress in small-model design and efficient inference, visual encoder remains the primary bottleneck for on-device deployment.

\textbf{Efficiency Improvements for On-device MLLMs.}
Improving the efficiency of multimodal large language models has attracted increasing attention, leading researchers to pursue optimization from model design to system-level deployment. To address the computational burden introduced by visual inputs, visual token compression has emerged as an important direction. For example, MiniCPM-V \citep{minicpmv-4.5} employs a Perceiver Resampler, which uses a single layer of cross-attention to compress hundreds of visual tokens into a small, fixed number of feature vectors, significantly reducing the sequence length processed by the language model. \cite{vico} introduces visual consistency learning, allowing the model to represent images with different numbers of visual tokens depending on their semantic complexity. Moreover, knowledge distillation \citep{llmkd1,llmkd2,vlmkd} has been further developed in multimodal settings, enabling a student model to absorb richer visual understanding and cross-modal alignment knowledge by mimicking the intermediate representations or outputs of a stronger teacher model. These algorithmic optimizations, combined with techniques such as model quantization (QAT/PTQ) \citep{liuspinquant,xiao2023smoothquant,liu2024llm}, operator fusion in inference frameworks, collectively form a comprehensive technical stack supporting efficient on-device multimodal inference.

%% file: sec/3_method.tex
\section{HyperVL}
\subsection{Model Architecture}
As illustrated in Figure \ref{img:model architecture}, HyperVL comprises four core modules: a visual resolution compressor, two visual encoders of different sizes, a vision-language projector, and a shared large language model (LLM). Detailed descriptions of each module are provided below:

\begin{figure*}[htp!]
\vskip -0.05in
\centering
\includegraphics[width=0.99\linewidth]{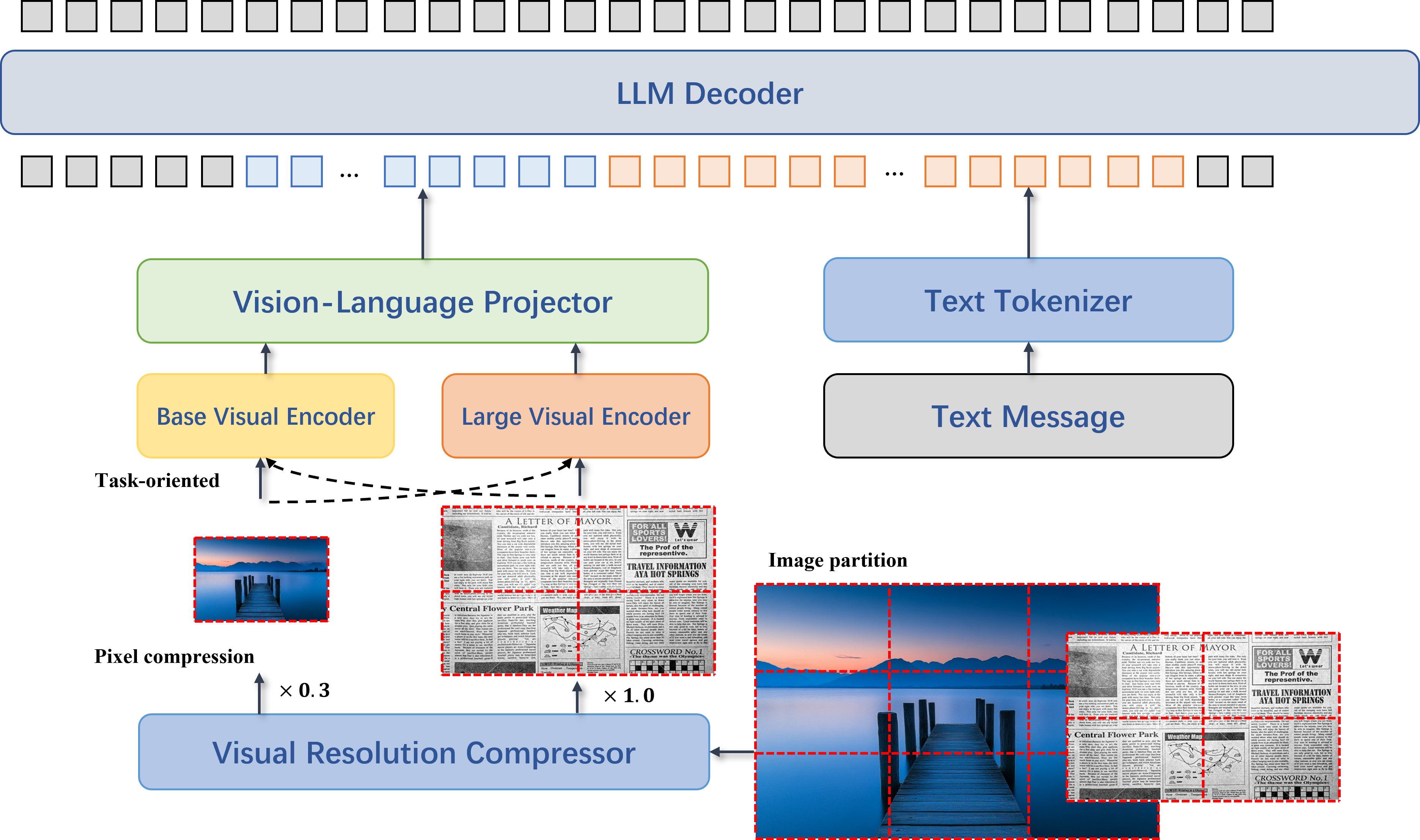}
\caption{The overall architecture of HyperVL mainly includes a visual resolution compressor, two visual encoders of different sizes, a vision-language projector, and a shared LLM.}
\label{img:model architecture}
\vskip -0.1 in
\end{figure*}

\textbf{Visual Resolution Compressor.} We observe that high-resolution images are not invariably required—larger images significantly increase the inference latency and memory footprint of on-device ViT, while in some scenarios, low-resolution images can already support high-quality responses. To address this, we design a visual resolution compressor that predicts the compression ratio based on the information density of the input image, scaling the image down to 10\%–100\% of its original size. The scaled image is then fed into the visual encoder.

\textbf{Visual Encoders.} We adopt two ViT variants of different sizes—SigLIP2-Large and SigLIP2-Base \citep{siglip2} as visual encoders, which consist of 24 layers (300M parameters) and 18 layers (93M parameters), respectively. To handle images with arbitrary aspect ratios and resolutions, we integrate the AnyRes \citep{anyres} approach: it preserves the original aspect ratio of the input image through minimal scaling and applies additional padding if necessary. The processed image is then tiled into non-overlapping patches. Both ViT encoders share the same LLM and can be flexibly switched between different downstream tasks.

\textbf{Vision-language Projector.} To map visual representations to the input space of the LLM, we employ a two-layer MLP as the projector, aligning the output of the ViT with the embedding layer of the LLM. To reduce the length of visual tokens fed into the LLM, we use pixel shuffle to compress the token length to 1/4 of its original size.

\textbf{Large Language Model (LLM).} Considering the balanced size and superior performance, we adopt the Qwen3 1.7B model \citep{qwen3} as the backbone LLM.

\begin{table}[htbp]
  \centering
  \caption{Configuration of HyperVL Model.}  % 表题：说明表格主题
  \label{tab:hyperlm_vl}  % 引用标签
  \begin{tabular}{l c l l}  % 列对齐：左、中、左、左
    \toprule[0.8pt]
    Model Name & \#Param (B) & Vision Encoder & Language Model \\
    \midrule[0.4pt]
    HyperVL & 1.8 & SigLIP2-Base & Qwen3-1.7B \\
    HyperVL ViT\textsubscript{L} & 2.0 & SigLIP2-Large & Qwen3-1.7B \\
    \bottomrule[0.8pt]
  \end{tabular}
\end{table}

\subsection{Dual Consistency Learning}
To accommodate the computational constraints of diverse on-device platforms, we design a dual-branch vision-language model architecture. The model incorporates two visual encoders of different sizes while sharing an single LLM. During on-device execution, the system can dynamically switch between branches based on task types or device capabilities, thereby balancing performance and resource consumption. 

Specifically, input images can be processed by either visual encoder to extract visual features, which are then projected into a semantic space aligned with textual representations before being fed into the language model for text generation or multimodal reasoning. This design aims to ensure that, whether using the small branch or the larger branch, the model can produce semantically consistent outputs with comparable performance. To this end, we employ a semantic consistency training strategy. Through an alternating dual-branch training mechanism, both visual encoders learn robust semantic mappings. In addition, a teacher–student distillation approach is applied to align the output distribution of the lightweight branch with that of the high-precision branch. 

\textbf{Dual-Branch Alternating Training.} 
The model is optimized using a step-wise alternating training strategy, in which two branches are activated alternately throughout training. This alternating optimization allows the model to learn a stable and consistent semantic mapping space under different visual representations.

\textbf{Semantic Consistency Distillation.} 
To further reduce the semantic distribution gap between the two branches, we introduce a Semantic Consistency Distillation mechanism. Specifically, the larger branch is treated as the teacher model, generating a teacher distribution $p_{\theta_l}(\cdot)$ for a given input. Simultaneously, the smaller branch serves as the student model, producing the prediction distribution $p_{\theta_b}(\cdot)$. By imposing a temperature-smoothed KL divergence constraint on the two distributions, the student branch gradually approximates the semantic representations of the teacher branch during training, achieving consistent cross-branch semantic alignment.
The loss function for this process is defined as:
% \begin{equation}
%     \mathcal{L}_{\text{total}} = (1 - \alpha)\,\mathcal{L}_{\text{task}} + \alpha\,\mathcal{L}_{\text{KD}}
% \end{equation}
% where $\alpha \in [0,1]$ is a weighting coefficient, \(\mathcal{L}_{\text{task}}\) is the task loss, and \(\mathcal{L}_{\text{KD}}\) is the distillation loss. The task loss \(\mathcal{L}_{\text{task}}\) adopts the standard causal language modeling loss, which calculates the cross-entropy between predicted logits and ground-truth labels. The distillation loss \(\mathcal{L}_{\text{KD}}\) uses token-level KL divergence, computed based on the teacher logits \(z_T\) and student logits \(z_S\):
% \begin{equation}
% \mathcal{L}_{\text{KD}} = \mathrm{KL}\!\left( \mathrm{Softmax}\!\left(\tfrac{z_T}{T}\right) \Big\|\, \mathrm{Softmax}\!\left(\tfrac{z_S}{T}\right) \right)
% \end{equation}
\begin{equation}
\mathcal{L}_{\text{KD}} = \mathrm{KL}\!\left( \pi_{\theta_s}\left(y_i|y_{<i},I\right) \Big\|\, \pi_{\theta_l}\left(y_i|y_{<i},I\right) \right)
\end{equation}
% where \(T\) is the distillation temperature, which controls the smoothness of the softmax distribution. 
Note that this loss is only computed on text tokens, skipping image tokens, to ensure the effectiveness and robustness of the distillation signal.

% \subsection{Dynamic Resolution Selection Module}
\subsection{Visual Resolution Compressor}
\begin{figure*}[htp!]
\vskip -0.05in
\centering
\includegraphics[width=0.99\linewidth]{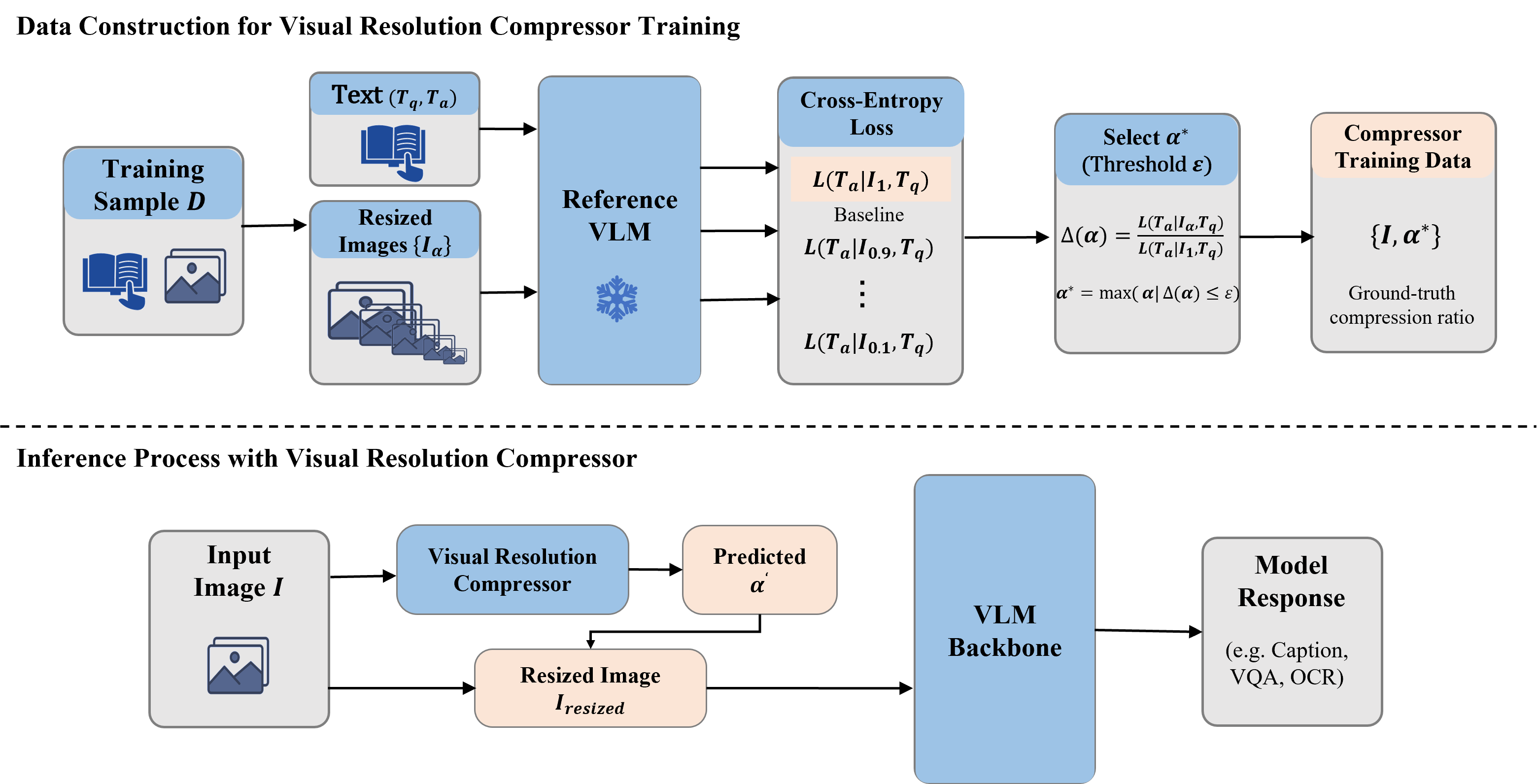}
\caption{Architecture of the Visual Resolution Compressor. The top panel illustrates the data construction pipeline used for training the compressor, while the lower shows the inference process. The bottom panel depicts the inference workflow, where the lightweight compressor dynamically predicts the optimal resolution to accelerate processing.}
\label{img:compressor}
\vskip -0.1 in
\end{figure*}

In multimodal large language models, input resolution governs not only the cost of visual encoding but also directly affects the computational complexity of LLM and the KV-cache memory footprint by determining the number of visual tokens. Higher resolutions capture more fine-grained visual information but incur substantially higher compute and memory costs during both training and inference. In contrast, tasks with low visual information density (e.g., object classification) can be accurately solved under lower input resolutions. Motivated by this, we propose a lightweight, plug-and-play Resolution Compressor designed to automatically predict the optimal compression ratio for input images. Its primary objective is to substantially reduce the number of visual tokens while preserving task performance, thereby lowering the overall computational overhead.

The design of the component follows three key principles: (1) On-device deployability: A lightweight architecture with constrained computation and storage overhead, compatible with mobile or embedded devices; (2) Plug-and-play capability: The module is independent of specific MLLM architectures and can be directly integrated into any pre-trained MLLM without additional architectural modifications; (3) Fine-grained controllability: Providing multi-level compression ratios to better balance computational overhead and performance. Compared with ViCO  \citep{vico}, which only supports two fixed compression ratios (0.25 and 1), the proposed visual resolution compressor enables finer-grained compression ratio selection (0.1–1.0), allowing more precise control over visual encoding resolution and computational cost.

\textbf{Data Construction.} 
To construct supervised data for training the compressor, we first calculate the ground-truth compression ratio for each image in the training samples. The design principle is to select the smallest compression ratio that does not introduce noticeable degradation. As illustrated in Figure \ref{img:compressor}, the specific procedure is as follows:

For each training sample \(D_i = \{I, T_p, T_a\}_i\) (where \(I\) denotes the image, \(T_p\) the prompt text, and \(T_a\) the ground-truth response), we compress the image \(I\) at multiple ratios to obtain a set of compressed variants \(\{I_{\alpha}\}\), where \(\alpha \in \{0.1, 0.2, \dots, 1.0\}\). Each image-text pair \(\{I_\alpha, T_p, T_a\}\) is then fed into a reference MLLM, from which we compute the cross-entropy loss. We measure the relative loss ratio by comparing each compressed loss against the loss of the uncompressed input. Specifically, using the baseline loss \(L(T_a|I_{1.0}, T_p)\), we define
% we compute the performance drop of \(L(T_a|I_\alpha, T_p)\) under different compression ratios and identify the maximum acceptable compression ratio \(\alpha^\star\) that the degradation does not exceed a predefined threshold \(\epsilon\):
\begin{equation}
\Delta(\alpha) = \frac{L(T_a|I_\alpha, T_p)}{L(T_a|I_{1.0}, T_p)}
\end{equation}
which quantifies the relative increase in loss caused by compression. We then identify the maximum acceptable compression ratio
\begin{equation}
\alpha^\star = \max\{\alpha \mid \Delta(\alpha) \le \epsilon\}
\end{equation}
where $\epsilon$ is a predefined loss tolerance threshold. This $\alpha^\star$ is used as the supervision signal for the compressor to predict the optimal compression ratio for each image.

% Here, \(\epsilon\) is the tolerance for performance loss, ensuring that the selected compression ratio does not significantly affect task performance in a statistical sense. This ratio serves as the supervision signal for the compressor to predict the optimal compression ratio for each image.

\textbf{Compressor Training.} 
To satisfy the computational and latency constraints of on-device deployment, the compressor is designed as a lightweight, low-parameter model. We adopt MobileNet \citep{mobilenet-v4} as the visual backbone to reduce computational complexity and memory footprint, while preserving task performance.
% Since the model requires fixed-resolution input (\(r \times r\)), we scale the original image and the maximum acceptable compression ratio proportionally:
As the model requires a fixed resolution of \(r \times r\), we adjust the original image and the corresponding maximum acceptable compression ratio accordingly:
\begin{equation}
    \alpha' = \frac{w \times h \times \alpha^\star}{r^2}
\end{equation}
where \(w\) and \(h\) represent the width and height of the original image, respectively. The training data \(\{I_{resize}, \alpha'\}\) is then prepared, where \(I_{resize}\) denotes the input image resized to \(r \times r\).

The resized image is passed through the visual backbone to extract features, followed by a learnable weighted pooling layer and a multi-layer perceptron (MLP) that predicts the scaling factor \(\hat{\alpha}'\).
The compressor is trained to predict the optimal scaling ratio of the input image, using the Mean Squared Error (MSE) as the loss function.
% \begin{equation}
%     \mathcal{L}_{\text{compressor}} = \frac{1}{N}\sum_{i=1}^N \big\| \hat{\alpha}'_i - \alpha'_i \big\|^2
% \end{equation}

% Here, \(\hat{\alpha}'_i\) denotes the predicted scaling factor for the $i$-th sample, and \(\alpha'_i\) is the corresponding target scaling ratio.

\textbf{Module Integration.} 
During inference, the compressor first predicts the normalized scaling factor \(\hat{\alpha}'\) from the resized input \(I_{resize}\) and then restores it to the corresponding compression ratio for the original image, as illustrated in Figure \ref{img:compressor}. 
This ratio is subsequently applied to the original high-resolution image. This process enables adaptive balancing of computational load and task performance on on-device platforms: for complex scenarios, the compressor preserves the original high-resolution image to maintain visual fidelity; for semantically simple or low-information-density scenarios, it selects a higher compression ratio to substantially reduce computational and memory overhead. 
Importantly, the compressor is agnostic to the MLLM backbone and can be integrated as a standalone plug-in into any model.

\subsection{Training Data} 
This section describes the construction of the multimodal training data used in this study. We first introduce the composition of the training corpus across different task categories, and then present the data pipeline used for data processing and organization.
\subsubsection{Overview}
To support the comprehensive capabilities of multimodal large models in visual understanding, cross-modal reasoning, and generation tasks, this study constructs a large-scale multimodal training corpus covering image caption, visual question answering (VQA), OCR, document understanding, grounding and counting, GUI interaction, STEM, and text-only tasks. 
The training corpus integrates a combination of open-source and synthetic data to balance task diversity and scenario generality. In particular, several large-scale open-source datasets—such as Infinity-MM \citep{infinity-mm}, FineVision \citep{finevision}, LLaVA-One-Vision-1.5 \citep{llavaonevision15}, LLaVAR \citep{llavar}, WIT \citep{WIT}, MMInstruct \citep{mminstruct}, ShareGPT4V \citep{sharegpt4v}, and ShareGPT4o \citep{sharegpt4o}—are comprehensive multimodal resources that span multiple task categories, including image captioning, VQA, instruction-following, and dialogue. These datasets provide rich visual and textual annotations, enabling the model to learn cross-modal understanding and reasoning in a variety of contexts. 
Throughout training, we dynamically adjust the proportions of different data modalities and task types across stages to balance training efficiency, model stability, and multi-task generalization. The specific data sources described below focus on task-specific datasets that complement the comprehensive multi-task datasets introduced above.

% \subsubsection{Pre-Training Data}
\textbf{Image Caption.} 
Image captioning data is used to train the model’s visual-to-linguistic generation ability, covering general scenarios and fine-grained descriptions. Task-specific datasets include Wukong \citep{gu2022wukong100millionlargescale}, Laion \citep{webster2023deduplicationlaion2b}, CoCo-Caption \citep{chen2015microsoftcococaptionsdata}, TextCap \citep{sidorov2020textcapsdatasetimagecaptioning}, and OpenImages-Caption \citep{openimages}. Notably, we re-annotate original images using captioning models—especially for the WIT dataset, where we leverage its original rich text as contextual information to incorporate more comprehensive world knowledge.

\textbf{Visual Question Answering (VQA).} 
VQA data is designed to improve the model’s capabilities in visual reasoning, knowledge-based QA, and multi-turn dialogue, encompassing open-domain, knowledge-driven, and complex reasoning tasks. These datasets exhibit significant diversity in question types, reasoning depth, and knowledge dependency, ensuring the model’s ability to transition from perception to cognition. 
Task-specific open-source datasets, such as GQA \citep{gqa}, TallyQA \citep{tallyqa}, A-OKVQA \citep{a-okvqa}, Objects365 \citep{objects365}, IconQA \citep{iconqa}, and VisDial \citep{visdial}, are used to provide examples of visual reasoning and dialogue scenarios.

\textbf{Optical Character Recognition (OCR).} 
OCR capability is a core foundational skill for MLLMs. To enhance the model’s ability to recognize and understand text in images, our data covers multiple languages and scenarios (natural scenes, documents, invoices, and tables). The data is highly diverse in font styles, languages, and scene complexity, supporting generalization in cross-lingual text understanding. Additionally, data augmentation (e.g., cropping, rotation, background replacement, and blurring) is applied to improve the model’s generalization and robustness.Representative open-source datasets include Laion-COCO \citep{laioncoco}, Wukong-OCR \citep{gu2022wukong100millionlargescale}, SynthDoG-EN \citep{synthdog-en}, SynthDoG-ZH \citep{synthdog-zh}, LSVT \citep{lsvt}, EATEN \citep{eaten}, and Chinese-Text-Recognition \citep{chinese-text-recognition}, providing annotated samples across multiple scripts and domains.

\textbf{Document Understanding.}
Document understanding data is used to train the model’s ability to parse, interpret, and reason over structured and unstructured documents, including forms, receipts, tables, and handwritten notes. The datasets cover a range of scenarios such as text layout analysis, entity recognition, and document-level question answering, aiming to improve both visual and semantic comprehension of documents.  
Selected open-source datasets, such as DUDE \citep{dude}, UniMER-1M \citep{unimer-1m}, and IAM-line \citep{iam-line}, provide representative examples of diverse document types and annotation styles, supporting the model’s generalization across structured and unstructured document understanding tasks.

\textbf{Grounding and Counting.} 
Grounding data is annotated with single or multiple bounding boxes. Basic datasets are selected from public sources such as Visual Genome \citep{visualgenomeconnectinglanguage}, RefCOCOs \citep{refcoco}, Flickr30k-Entities \citep{flickr30k}, and Groma \citep{groma}. These datasets are filtered and categorized into four types: object referencing, regional captioning, referred entity recognition, and grounded image captioning. We design multiple prompt templates for model instruction and unify all coordinates into a relative coordinate system of [0, 1000].

\textbf{Graphical User Interface (GUI) Understanding.}
To enhance models’ understanding and navigation capabilities in real-world GUI environments, we construct a large-scale, multi-scenario GUI dataset covering key tasks such as icon recognition, interface element analysis, region localization, interaction reasoning, and agent action execution. The data are sourced from open-source UI icon libraries, real-world multi-platform usage scenarios, and automated task collection, encompassing a variety of interface types, including mobile, web, and desktop applications. Representative open-source datasets include AITW \citep{aitw}, AndroidControl \citep{androidcontrol}, AutoGUI \citep{autogui}, RicoSCA \citep{ricosca}, and WaveUI \citep{waveui}, providing diverse examples of GUI layouts and interaction patterns. By aligning icons with their functional semantics, the dataset enables models to learn mappings between visual appearance features of controls and interaction intents, gradually extending from element recognition to comprehensive page structure understanding and interaction reasoning.

\textbf{STEM (Science, Technology, Engineering, Mathematics).} 
STEM data is used for scientific reasoning, mathematical formula understanding, and professional knowledge QA, covering disciplines such as mathematics, physics, finance, medicine, and chemistry. It aims to strengthen the model’s logical reasoning and knowledge-based QA capabilities. In our corpus, we incorporate selected open-source STEM datasets, for example ART500K \citep{art500k}, BIOMEDICA \citep{biomedica}, and ScienceQA \citep{scienceqa}, which offer a variety of domain-specific questions and reasoning tasks.

\textbf{Text-Only Instruction Data.} 
To maintain the text capabilities of the multimodal model, we incorporate a certain proportion of text pre-training data and instruction-following data. Model performance changes are monitored through instruction-based evaluations.

\subsubsection{Data Pipeline}
\begin{figure*}[htp!]
\vskip -0.05in
\centering
\includegraphics[width=0.99\linewidth]{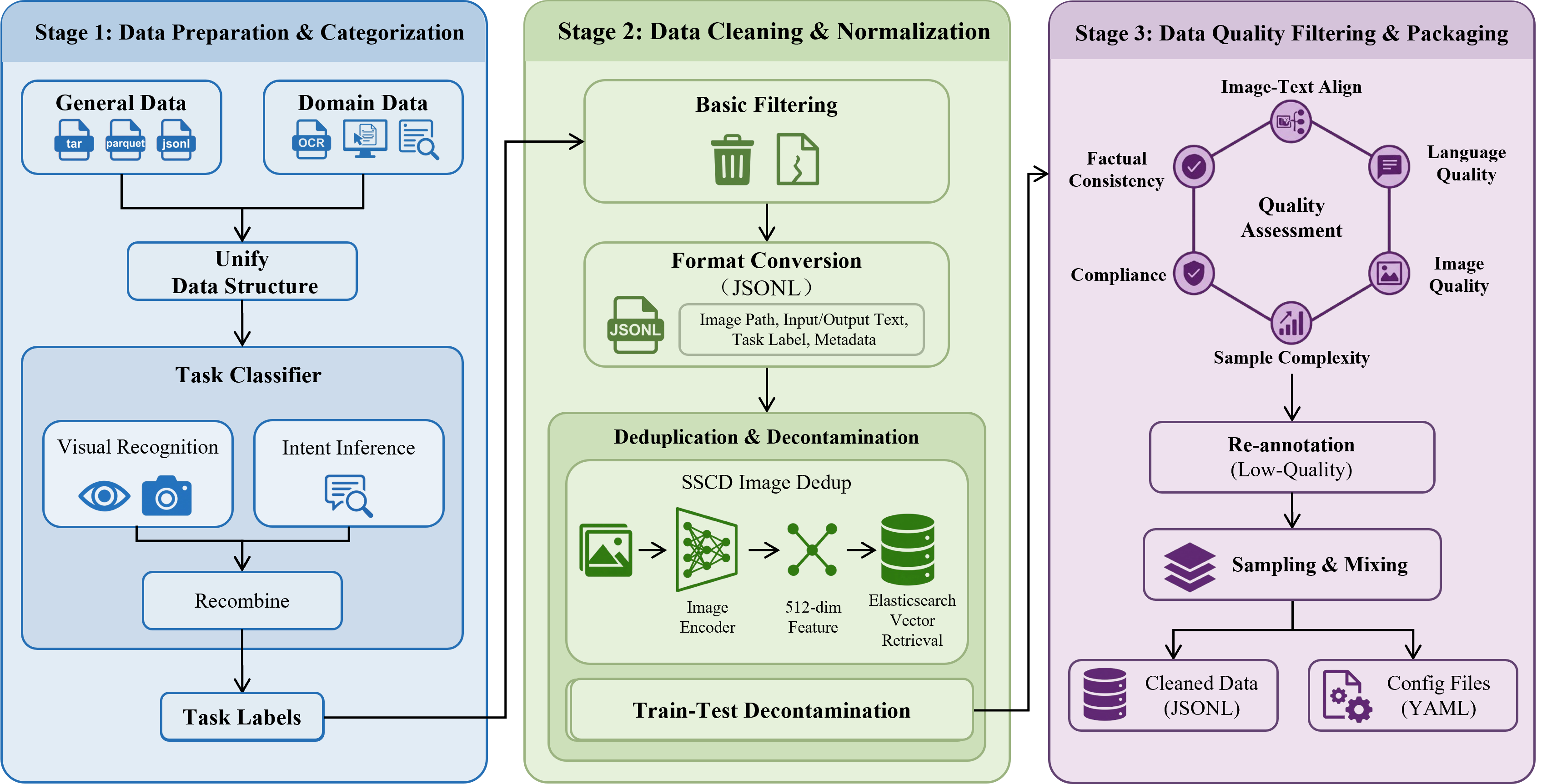}
\caption{\textbf{Data pipeline.} We adopt a three-stage data governance pipeline consisting of data preparation and categorization, data cleaning and normalization, and data quality filtering and mixed packaging, enabling large-scale construction of high-quality multimodal training samples with strong controllability and consistency.}
\label{img:framework}
\vskip -0.1 in
\end{figure*}

To facilitate the construction of large-scale, high-quality multimodal training samples, HyperVL designs and implements a systematic data governance pipeline that spans data collection, cleaning, filtering, categorization, and deduplication. As illustrated in Figure \ref{img:framework}, the pipeline is organized into three stages: (i) data preparation and categorization, (ii) data cleaning and normalization, and (iii) data quality filtering and mixed packaging.

\textbf{Data Preparation and Categorization.} 
The HyperVL multimodal corpus comprises both general task data and domain-specific task data, stored in standardized formats such as tar, parquet, json, and jsonl. General data typically contains heterogeneous task types and requires automatic task partitioning after decompression and structural unification. To enable fine-grained sampling control and improve data controllability, we develop a detailed task categorization system using a task decomposition–recombination strategy, in which categorization is decomposed into two parallel subtasks: (i) content-based visual recognition and (ii) query-driven intent inference from textual instructions, with the final task label obtained through logical recombination of the two predictions. This design achieves a stable categorization accuracy exceeding 96\% on complex image–text samples, substantially improving sampling controllability and distribution balance in downstream training. In contrast, domain-specific data (e.g., OCR and GUI grounding) is inherently organized by task type and can be directly forwarded to subsequent processing stages.

\textbf{Data Cleaning and Normalization.} 
This stage aims to transform raw multimodal corpora collected from diverse sources into unified samples. We first perform automated quality inspection and remove samples with corrupted or blurred images, missing or misaligned text, incomplete question–answer pairs, trailing duplication or gibberish, and abnormal or inconsistent fields, thereby ensuring structural integrity and semantic consistency.

Next, all remaining samples are converted into a unified HyperVL JSONL format, retaining essential fields such as image paths, input text, output text, task labels, and metadata to ensure cross-dataset consistency. Dual-modal deduplication is conducted over the entire corpus, followed by explicit training–test decontamination to prevent evaluation data leakage.

For image collections exceeding billions of samples and exhibiting high source heterogeneity, we further deploy an efficient image deduplication system based on Self-Supervised Copy Detection (SSCD) \citep{sscd}. Unlike traditional perceptual hashing methods such as pHash \citep{phash}, SSCD encodes each image into a 512-dimensional deep feature representation and supports large-scale similarity search via Elasticsearch-based vector retrieval. This feature-level matching enables robust detection of semantic duplicates under cross-dataset, cross-style, and cross-resolution variations, significantly improving deduplication accuracy and robustness.

\textbf{Data Quality Filtering and Mixed Packaging.} 
On the cleaned and deduplicated corpus, HyperVL conducts a multi-dimensional data quality assessment followed by automated refinement. We design a unified quality evaluation framework spanning six primary dimensions—image–text alignment, linguistic quality, image quality, factual consistency, sample complexity, and compliance—resulting in twelve quantifiable sub-indicators. The evaluation procedure integrates metric normalization with type-adaptive weighting, assigning importance to indicators across data categories, and performs a joint assessment via rule-based heuristics and the Qwen3-VL model, ensuring both robustness and flexibility. Samples identified as severely low quality are automatically re-annotated in an iterative manner to improve overall data reliability. Finally, HyperVL applies hierarchical sampling and weighted mixing based on task categories and quality tiers, producing structured clean JSONL training sets and corresponding YAML configuration files for multi-stage training.

\subsection{Training Pipeline}
% \subsubsection{Pre-training}
We divide the pre-training of the multimodal model into three phases: the vision-language alignment phase, the knowledge enhancement phase, and the multi-task training phase. The parameter configurations and data details for each phase are presented in Table \ref{tab:hyperlm_vl_training}.

\begin{table}[htbp]
\centering
\caption{Overview of HyperVL training stages.}
\label{tab:hyperlm_vl_training}
% \renewcommand{\arraystretch}{1.1} % 调整行高以获得更好的视觉效果
% \resizebox{\textwidth}{!}{%
\begin{tabular}{cccc}
\toprule
\textbf{Stages} 
  & \textbf{Alignment} 
  & \textbf{Knowledge Enhancement} 
  & \textbf{Multi-Task Training} \\
\midrule

Dataset &
\makecell[c]{Caption} &
\makecell[c]{Caption \\ + VQA \\ + OCR \\ + Interleaved image \\ + Grounding \\ + STEM \\ + Text} &
\makecell[c]{Multitask data} \\
\midrule
Tokens & 82.8B & 214.6B & 55.1B \\
Sequence Length & 4096 & 8192 & 8192 \\
Learning Rate & 5e-4 & 5e-5 & 2e-5 \\
Trainable Parameters & \makecell[c]{Adapter} 
  & \makecell[c]{ViT + Adapter + LLM} 
  & \makecell[c]{ViT + Adapter + LLM} \\
\bottomrule
\end{tabular}
% }
\end{table}

\textbf{Stage 1: Vision-Language Alignment.} 
The visual encoders adopt parameters from two variants of SigLIP2 (Base and Large). Compared to larger alternatives such as Qwen2VL-ViT 675M and InternVL 300M, their structure and size are more suitable for mobile deployment. During this stage, the parameters of the ViT and LLM are frozen, while the adapter parameters are randomly initialized. We train the adapter using image-text captioning data with a learning rate of 5e-4. The primary goal of this stage is to align the visual encoder with the language model.

\textbf{Stage 2: Knowledge Enhancement.} 
The second stage involves joint vision-language pre-training. We unfreeze the parameters of the ViT and LLM and perform full-parameter pre-training with a relatively low learning rate. In this stage, the training data includes multi-type, multi-task image-text data and text-only data, enabling the model to acquire extensive visual knowledge. During pre-training on this data, we only compute the loss for text tokens while excluding the loss for image tokens. The maximum sequence length used in this stage is 8192.

\textbf{Stage 3: Multi-Task Training.} 
To further enhance the model’s general capabilities and reasoning ability, we select high-quality multi-task data, including STEM, mathematics, reasoning, and synthetic data. Existing studies have shown that high-quality reasoning-oriented pre-training datasets can moderately improve general reasoning performance, while synthetic data yields significant benefits across various downstream tasks. The synthetic data here includes short Chain-of-Thought (CoT) and long CoT sequences, covering both text and multimodal reasoning tasks. In this stage, the maximum sequence length is increased to 8192, and the model is trained on approximately 23M high-quality data. Ultimately, this stage enhances the model’s ability to handle complex, multi-step reasoning processes.

% \subsubsection{Post-training}
% A core advantage of reinforcement learning (RL) lies in its ability to incorporate negative samples, thereby pruning low-quality regions in the model’s output space and improving overall response quality. We introduce RL in the post-training process, which consists of two main phases: Preference Alignment (MPO) and Reinforcement Learning (RL). Specifically, Mixed Preference Optimization (MPO) [] is used to enhance the model’s instruction understanding and align with preference data, while RL based on Group Relative Policy Optimization (GRPO) [] can further strengthen the model’s reasoning capabilities.

% \textbf{Mixed Preference Optimization (MPO)}
% Direct Preference Optimization (DPO) has emerged as a primary method for aligning LLMs with human preferences. It avoids complex policy RL pipelines and is suitable for training non-chain-of-thought models. Leveraging its effectiveness in natural language processing, recent studies have extended the application of DPO to multimodal scenarios []. A novel approach called Mixed Preference Optimization (MPO) has been proposed, which demonstrates improved performance across various multimodal reasoning evaluation benchmarks. We adopt the MMPR dataset and MPO method from [228]. During training, a joint loss function is used, consisting of preference loss 

% \textbf{Group Relative Policy Optimization (GRPO)}

%% file: sec/4_experiment.tex
\section{Evaluation}
\subsection{Evaluation Setting}
To comprehensively evaluate the model’s performance in multimodal understanding, reasoning, and generation, we conduct a systematic evaluation on both public and internal benchmarks. Our model is compared against state-of-the-art multimodal baselines with comparable parameter scales to ensure fair and meaningful comparisons. For public benchmarks, we adopt the OpenCompass open-source evaluation framework, VLMEvalKit, which provides a standardized and reproducible evaluation protocol. GPT-4o is employed as the judge model for automated answer evaluation and consistency checking, ensuring the objectivity and reliability of the reported results. %Detailed descriptions of these benchmarks are provided in Tables and in the Appendix.

\subsection{Public Benchmarks}
The public benchmark evaluation framework encompasses four core leaderboards, spanning multimodal reasoning, OCR and document understanding, comprehensive multimodal capabilities, and hallucination detection. The evaluated benchmarks are detailed as follows:

\subsubsection{Multimodal Reasoning and Mathematics} Models are evaluated for cross-modal reasoning and mathematical problem-solving capabilities under this leaderboard, including:

\textbf{MMMU\_val} \citep{mmmu} A multimodal expert-level reasoning benchmark covering over 30 disciplines such as science, engineering, medicine, and finance, emphasizing cross-modal knowledge integration and professional-level visual understanding.

\textbf{MMMU\_Pro\_V} \citep{mmmu-pro} A robust extension of MMMU featuring vision-only question embedding, enhanced option diversity, and strict filtering to eliminate text-only solvable items.

\textbf{MathVista} \citep{lu2024mathvistaevaluatingmathematicalreasoning} A small-scale benchmark for visual mathematics, focusing on formula understanding, geometric reasoning, and reasoning in mixed text-image mathematical scenarios.

\subsubsection{OCR, Chart, and Document Understanding} In this leaderboard, models are evaluated on text recognition, chart parsing, and structured document understanding, including:

\textbf{AI2D} \citep{ai2d} A diagram reasoning dataset based on educational scenarios, testing models’ understanding of explanatory images and conceptual relationships through flowcharts and structure diagrams.

\textbf{ChartQA} \citep{chartqa} Evaluates models’ understanding and reasoning over visual charts, including bar charts, line charts, and pie charts, with emphasis on structured information extraction and numerical reasoning.

\textbf{TextVQA} \citep{textvqa} Focuses on recognition and question answering for text embedded in visual scenes, testing OCR and semantic understanding together.

\textbf{DocVQA} \citep{docvqa} A visual question answering benchmark on scanned documents and PDFs, assessing structured document understanding and information extraction.

\textbf{InfoVQA} \citep{infographicvqa} Evaluates document information extraction and QA in highly structured scenarios such as financial statements, invoices, and contracts.

\textbf{OCRBench} \citep{ocrbench} A general OCR benchmark covering natural scenes, scanned documents, and receipts, focusing on robustness to complex fonts, multilingual text, and noisy environments.

\textbf{ChineseOCRBench} \citep{ChineseOCRBench} A Chinese-focused high-difficulty OCR benchmark, covering multiple fonts, diverse scenarios, complex layouts, and handwritten text.

\textbf{SEEDBench} \citep{li2023seedbenchbenchmarkingmultimodalllms} A large-scale multimodal benchmark featuring 19K multiple-choice questions across 12 spatial and temporal understanding dimensions, designed for rigorous evaluation of generative multimodal large language models.

\subsubsection{Comprehensive Multimodal tasks} This evaluation setting measures model performance across integrated multimodal tasks, including:

\textbf{MME} \citep{mme} A comprehensive benchmark comprising 14 perception and cognition subtasks—measuring vision-language understanding through manually crafted, concise instruction-answer pairs to enable fair and quantitative evaluation of multimodal large language models.

\textbf{MMBench\_CN} \citep{mmbench} A comprehensive benchmark featuring diverse, quality-controlled questions, rigorous evaluation strategies, and standardized multiple-choice formats for robust assessment of multimodal large language models.

\textbf{MMBench v1.1 (EN/CN)} \citep{mmbench} A comprehensive multimodal benchmark assessing image understanding, description, reasoning, and multilingual question answering.

\textbf{MMVet} \citep{mmvetevaluatinglargemultimodal} A visual reasoning benchmark emphasizing evidence consistency and factual accuracy, testing models’ ability to avoid hallucinations across cross-image facts and visual details.

\textbf{MMStar} \citep{mmstar} A multimodal instruction-following benchmark for complex visual scenarios, covering reasoning, localization, relationship understanding, and multi-step chain-of-task reasoning.

\textbf{MMMB\_CN} \citep{mmmb} A multilingual multimodal benchmark comprising 12,000 multiple-choice questions across 15 categories and 6 languages, designed to rigorously evaluate cross-lingual vision–language understanding in multimodal large language models. In this work, we utilize the MMMB\_CN subset to specifically assess the Chinese language capability of our model.

\textbf{MME-RealWorld\_CN} \citep{zhang2025mmerealworldmultimodalllmchallenge} A high-resolution multimodal benchmark designed to evaluate multimodal large language models in complex real-world scenarios, featuring diverse tasks that challenge fine-grained perception, reasoning, and robustness under visually demanding conditions.

% \textbf{CCBench} \citep{mmbench} A multimodal benchmark for evaluating LLMs on Chinese cultural content, covering tasks such as QA, image description, and VQA with 510 Chinese multiple-choice questions.

\subsubsection{Hallucination Evaluation} Model tendencies toward visual and factual hallucinations are examined through the following benchmarks:

\textbf{HallusionBench} \citep{hallusionbench} Diagnoses factual hallucinations in visual scenes, focusing on unsupported inferences and incorrect visual details.

\textbf{POPE} \citep{pope} Tests robustness in determining object existence through contrastive question-answering, measuring visual hallucination rates.

\textbf{CRPE} \citep{crpe} Evaluates factual consistency and reasoning fidelity in complex visual-text scenarios, particularly for content fabrication in reasoning chains.
% \textbf{GUI Task Benchmarks.}

\subsection{Internal Benchmarks}
While public benchmarks provide a standardized assessment of general capabilities, they often fail to capture the complexities encountered in real-world deployments. To evaluate the practical utility of our model, we construct a specialized benchmark tailored to internal scenarios, encompassing four core tasks: user intent recognition and recommendation, image-text creation, UI understanding and structured parsing, and image relevance ranking.

\subsubsection{Intent Recognition and Recommendation} 
The benchmark evaluates a multimodal model’s ability to parse diverse visual information from device screenshots—including UI structures, text, and interaction states—to generate concise search queries that reflect the user's potential intent. Evaluation focuses on deep semantic understanding and latent intent extraction, which underpin downstream features such as proactive search recommendations. The assessment measures the model’s proficiency in interpreting UI visual structures, extracting search intent, and producing relevant, actionable queries under strict constraints. 
A rigorous evaluation protocol is employed: a generated query is considered successful only if it simultaneously satisfies six criteria—topic relevance, factual accuracy, logical soundness, linguistic fluency, recommendation value, and safety compliance. Failure in any single criterion results in overall rejection, ensuring high standards for deployment readiness.

\subsubsection{Image-Text Creation} 
Model performance is assessed on the ability to generate high-quality text closely aligned with both the application context (e.g., social media editors, review interfaces) and user-uploaded images. The task covers common content-generation and review-related application pages, including posting interfaces on platforms such as Xiaohongshu, WeChat, Weibo, Douyin, Kuaishou, and Dianping, as well as product review pages on platforms such as Dianping and Taobao. Performance is evaluated using a four-dimensional framework comprising topic relevance, multimodal alignment, content quality, and style consistency, ensuring that the generated text is semantically accurate, visually grounded, and strictly adherent to user-defined stylistic constraints.

\subsubsection{UI Understanding and Structured Parsing} 
The benchmark assesses the model’s capability to comprehend complex mobile interfaces and structurally extract key information, with a focus on order detail pages. Given a raw screenshot, the model performs joint analysis of textual content, visual layout, and semantic relationships to identify critical fields without reliance on predefined templates. Structured outputs are produced for downstream applications.

Performance is evaluated using Field Accuracy, defined as the ratio of correctly extracted fields to the total number of valid fields across the dataset:
\begin{equation}
\text{FieldAccuracy} = 
\frac{\sum_{i=1}^{N} C_i}{\sum_{i=1}^{N} M_i}    
\end{equation}
where $C_i$ represents the number of correctly extracted fields in sample $i$, and $M_i$ represents the total number of ground-truth valid fields in that sample.

\subsubsection{Image Relevance Ranking} 
This task employs a multimodal large model to perform fine-grained relevance assessment on candidate images retrieved during the initial coarse-retrieval stage. The primary objective is to filter out irrelevant or low-quality results and prioritize the most semantically pertinent images, thereby maximizing the overall precision of the final search results.
The model accepts a user query and a set of $N$ candidate images as input. For each (Query, Image) pair, the model performs deep cross-modal interaction modeling to compute a fine-grained relevance score. The candidate set is then re-ranked based on these scores to produce an optimized retrieval list.

Evaluation relies on Precision@10, Precision, and the F1 Score. Precision@10 measures the proportion of relevant images within the top-10 results of the re-ranked list:
\begin{equation}
    \text{Precision}@10 = \frac{|\text{Relevant Results in Top-K}|}{10}
\end{equation}
Precision evaluates the overall accuracy of relevance predictions (treating the task as a binary classification of relevant vs. irrelevant), calculated as:
\begin{equation}
    \text{Precision} = \frac{\text{True Positives (TP)}}{\text{True Positives (TP)} + \text{False Positives (FP)}}
\end{equation}
F1 Score represents the harmonic mean of Precision and Recall, providing a balanced view of the model's retrieval performance:
\begin{equation}
    \text{F1} = 2 \times \frac{\text{Precision} \times \text{Recall}}{\text{Precision} + \text{Recall}}
\end{equation}

\subsection{Experimental Results}
\subsubsection{Public Benchmarks}

% \usepackage{multirow}
% \usepackage{graphicx}
\begin{table}[]
\centering
\resizebox{\textwidth}{!}{%
\begin{tabular}{l|cccccccc|cc}
\toprule
\multicolumn{1}{l}{} &
  \multicolumn{1}{c}{} &
  \multicolumn{1}{c}{} &
  \multicolumn{1}{c}{} &
  \multicolumn{1}{c}{} &
  \multicolumn{1}{c}{} &
  \multicolumn{1}{c}{} &
  \multicolumn{1}{c}{} &
  \multicolumn{1}{c}{} &
  \multicolumn{1}{c}{} &
  \multicolumn{1}{c}{} \\
\multicolumn{1}{l}{\multirow{-2}{*}{\textbf{Benchmarks}}} &
  \multicolumn{1}{c}{\multirow{-2}{*}{\begin{tabular}[c]{@{}c@{}}\textbf{Qwen2-VL}\\ 2.2B\end{tabular}}} &
  \multicolumn{1}{c}{\multirow{-2}{*}{\begin{tabular}[c]{@{}c@{}}\textbf{Qwen3-VL}\\ 2.1B\end{tabular}}} &
  \multicolumn{1}{c}{\multirow{-2}{*}{\begin{tabular}[c]{@{}c@{}}\textbf{Ovis2}\\ 2.5B\end{tabular}}} &
  \multicolumn{1}{c}{\multirow{-2}{*}{\begin{tabular}[c]{@{}c@{}}\textbf{InternVL3}\\ 2.1B\end{tabular}}} &
  \multicolumn{1}{c}{\multirow{-2}{*}{\begin{tabular}[c]{@{}c@{}}\textbf{InternVL3.5}\\ 2.3B\end{tabular}}} &
  \multicolumn{1}{c}{\multirow{-2}{*}{\begin{tabular}[c]{@{}c@{}}\textbf{SAIL-VL1.5}\\ 2.5B\end{tabular}}} &
  \multicolumn{1}{c}{\multirow{-2}{*}{\begin{tabular}[c]{@{}c@{}}\textbf{SAIL-VL2}\\ 2.7B\end{tabular}}} &
  \multicolumn{1}{c}{\multirow{-2}{*}{\begin{tabular}[c]{@{}c@{}}\textbf{AndesVL}\\ 2.4B\end{tabular}}} &
  \multicolumn{1}{c}{\multirow{-2}{*}{\begin{tabular}[c]{@{}c@{}}\textbf{HyperVL}\\ 1.8B\end{tabular}}} &
  \multicolumn{1}{c}{\multirow{-2}{*}{\begin{tabular}[c]{@{}c@{}}\textbf{HyperVL}\\ ViT\textsubscript{L} 2.0B\end{tabular}}} \\
\midrule
\rowcolor[gray]{0.9} \multicolumn{11}{c}{\textit{Opencompass Scores}}         \\
\midrule
Opencompass\textsubscript{avg}  & 56.4  & 63.5    & 62.9    & 63.9   & 69.1   & 66.9    & 69.1    & 64.6   & 64.5   & 66.1   \\
\midrule
\rowcolor[gray]{0.9} \multicolumn{11}{c}{\textit{Multimodal Reasoning and Mathematics}}                              \\
\midrule
MMMU              & 41.1  & 53.4    & 41.4*   & 48.6   & 59.0   & 42.9    & 47.7    & 46.1   & 43.4   & 44.6   \\
MMMU\textsubscript{Pro\textsubscript{V}}      & 19.9*  & 28.1*    & 18.5*   & 17.5*  & 23.6*  & 16.4*    & 22.2*   & 22.0*   & 23.9   & 23.8   \\
MathVista         & 43    & 61.3    & 64*     & 57.0   & 71.8   & 67.3    & 71.1    & 64.9   & 66.2   & 65.9   \\
\midrule
\rowcolor[gray]{0.9} \multicolumn{11}{c}{\textit{OCR, Chart, and Document Understanding}}                            \\
\midrule
AI2D\textsubscript{TEST}        & 74.7  & 76.9    & 79.7*   & 78.7   & 78.8   & 83.7    & 83.0    & 80.1   & 81.8   & 83.1   \\
ChartQA\textsubscript{TEST}     & 73.5  & 77.6*   & 79.5*   & 80.2   & 80.7   & 84.8    & 85.4*   & 87.4   & 83.8   & 85.7   \\
TextVQA\textsubscript{VAL}      & 79.7  & 79.8*   & 78.1*   & 77.0   & 76.5   & 82.0    & 82.1*   & 79.9   & 75.7   & 78.8   \\
DocVQA            & 90.1  & 93.3    & 89.6*   & 88.3   & 89.4   & 91.6    & 93.1    & 94.2   & 91.3   & 92.2   \\
InfoVQA           & 65.5  & 72.4    & 69.7*   & 66.1   & 70.8   & 71.8    & 73.5*   & 74.2   & 70.2   & 74.6   \\
OCRBench          & 809   & 858     & 868*    & 835    & 836    & 885     & 895     & 846    & 830    & 859    \\
ChineseOCRBench   & 77.5* & 78.0*   & 81.8*   & 76.7*  & 66.2*  & 63.0*   & 76.2*   & 88.3*  & 80.4   & 84.3   \\
SEEDBench\textsubscript{IMG}    & 71.6* & 74.8*   & 73.3*   & 74.6*  & 63.8*  & 74.2*   & 77.3*   & 73.5*  & 74.1   & 75.3   \\
\midrule
\rowcolor[gray]{0.9} \multicolumn{11}{c}{\textit{Comprehensive Multimodal Tasks}}                                    \\
\midrule
MME               & 1872  & 2004.2* & 1406.7* & 2221.2 & 2123.0 & 1837.9* & 2156.1* & 2081.0 & 2070.7 & 2114.7 \\
MMBench\_DEV\textsubscript{CN}  & 73.5  & 76.28   & 72.2*   & 78.4   & 45.5*  & 71.7*   & 81.1*   & 77.8*  & 79.2   & 79.64   \\
MMBench\_V1.1\textsubscript{EN} & 72.2  & 77.8    & 71.7*   & 78.0   & 76.6   & 78.5    & 77.0    & 77.3   & 76.5   & 77.4   \\
MMBench\_V1.1\textsubscript{CN} & 72.2  & 75.5*   & 72.1*   & 78.6   & 44.4*  & 70.8*   & 80.1*   & 77.3   & 76.0   & 77.1   \\
MM-Vet            & 49.5  & 43.3    & 44.4*   & 62.2   & 71.7   & 61.4    & 68.7    & 52.0   & 56.3   & 59.0   \\
MMStar            & 48    & 58.3    & 54.7*   & 60.7   & 62.7   & 62.8    & 64.1    & 60.0   & 59.2   & 61.1   \\
MMMB\textsubscript{cn}          & 74.2  & 78.0*   & 78.9*   & 78.3   & 77.7   & 79.3*   & 82.6*   & 78.6*  & 80.5   & 79.3   \\
MME-RealWorld\textsubscript{CN}  & 44.3* & 55.2*   & 52.3*   & 47.5*  & 35.9*  & 47.9*   & 59.5*   & 56.7*  & 54.4   & 52.9   \\
\midrule
\rowcolor[gray]{0.9} \multicolumn{11}{c}{\textit{Hallucination Evaluation}}                                          \\
\midrule
HallBench         & 41.7  & 51.4    & 60.8*   & 42.5   & 48.6   & 49.8    & 51.7    & 51.8   & 49.2   & 51.5   \\
CRPE              & 68.3* & 73.7*   & 64.9*   & 71.5   & 75.6   & 61.1*   & 75.2    & 73.0   & 64.5   & 62.7   \\
POPE              & 86.4* & 88.8*   & 87.5*   & 89.6   & 87.2   & 84.5*   & 86.7*   & 87.9   & 87.7   & 88.9  \\
\bottomrule
\end{tabular}%
}
\caption{Comparison of HyperVL with other models with similar parameter scales on diverse benchmarks, which are grouped into 4 domains. Data marked with * are from our evaluation, while others are from their original papers or the OpenCompass leaderboard.}% The best results are marked in bold.}
\label{tbl:main}
\end{table}

Table \ref{tbl:main} presents a comprehensive comparison between HyperVL and representative vision–language models with similar parameter scales. The two rightmost columns report the performance of HyperVL and HyperVL ViT\textsubscript{L}, both trained under the same architectural framework.

Despite its compact scale, HyperVL achieves an OpenCompass average score of 64.5, which is competitive with several 2B–3B models. Notably, it demonstrates consistent strengths in OCR, chart, and document understanding tasks, achieving 83.8 on ChartQA, 81.8 on AI2D, and 91.3 on DocVQA. These benchmarks emphasize the model’s ability to capture fine-grained visual details and structured visual information, which are critical for real-world multimodal applications.

Scaling the model to HyperVL ViT\textsubscript{L} leads to stable and consistent improvements across most benchmarks, raising the OpenCompass average score to 66.1. Performance gains are observed not only in OCR-related tasks but also in comprehensive multimodal benchmarks such as MME and MMBench, indicating that the proposed framework scales smoothly with increased capacity.

In the multimodal reasoning and mathematics domain, HyperVL achieves a MathVista score of 66.2, outperforming some models of similar scale and remaining competitive with larger counterparts. On more challenging benchmarks such as MMMU, HyperVL maintains solid baseline performance, suggesting balanced reasoning capability despite the limited parameter budget.

Furthermore, both HyperVL variants exhibit stable hallucination-related performance on HallBench, with scores comparable to models of larger scale. This indicates that reducing model size does not lead to a noticeable degradation in reliability or safety-related behavior.

Overall, the results demonstrate that HyperVL offers a well-balanced trade-off between model capacity and performance. The base variant emphasizes parameter efficiency and strong visual understanding, while the large variant further improves general multimodal capability with minimal additional cost. These characteristics make HyperVL a practical and effective solution for lightweight yet general-purpose multimodal modeling.

\subsubsection{Internal Benchmarks}
\begin{table*}[htbp]
\renewcommand{\arraystretch}{1.1} % 调整行高以获得更好的视觉效果
\resizebox{\textwidth}{!}{%
\begin{tabular}{c|ccccccc|c}
\toprule
\multirow{2}{*}{\textbf{Benchmarks}} & \textbf{Qwen2-VL} & \textbf{Qwen3-VL} & \textbf{Ovis2} & \textbf{InternVL3} & \textbf{InternVL3.5} & \textbf{SAIL-VL2} & \textbf{AndesVL} & \textbf{HyperVL}\\
 & \textbf{2.2B} & \textbf{2.1B} & \textbf{2.5B} & \textbf{2.1B} & \textbf{2.3B} & \textbf{2.7B} & \textbf{2.4B} & \textbf{1.8B}\\
\midrule
% \rowcolor[gray]{0.9} \multicolumn{9}{c}{\textit{Multimodal Tasks}} \\
\begin{tabular}[c]{@{}c@{}}UI Understanding \\ and Structure Parsing \end{tabular} & 80.1 & \underline{91.7} & 78.1 & 83.5 & 70.6 & \textbf{92.7} & 86.7 & 84.2 \\
\midrule
\begin{tabular}[c]{@{}c@{}} Image Relevance \\ Ranking\end{tabular}   & 48.6 & 48.6 & 44.1 & 46.2 & 44.3 & \underline{50.5} & 49.4 & \textbf{51.5} \\
\midrule
% \begin{tabular}[c]{@{}c@{}}Grounding\\ \end{tabular}     &     &     &    &     &      &       &     &    \\
% \midrule
\begin{tabular}[c]{@{}c@{}} Intent Recognition \\ and Recommendation \end{tabular}   & 92.7 & 91.7 & 91.7 & 90.3 & 80.3 & 89.3 & \textbf{95.0}    & \underline{94.0} \\
\midrule
\begin{tabular}[c]{@{}c@{}} Image-Text \\ Creation \end{tabular}     & 21.2 & 28.9 & 37.7 & 45.4 & 43.9 & 38.6 & \underline{46.1} & \textbf{49.8} \\
% \midrule
% \rowcolor[gray]{0.9} \multicolumn{9}{c}{\textit{Text Tasks}} \\
% \begin{tabular}[c]{@{}c@{}}Message Information\\ Extraction \end{tabular}  & 69.33   & 84.83   & 83.81     & 87.46  & 33.30    & 60.09   & 81.30   &  82.55    \\
% \midrule
% \begin{tabular}[c]{@{}c@{}}Proofread \end{tabular}   & 36.96 & 67.39   & 86.95    & 81.16   & 50      &  -   &  -   &  54.34   \\
\bottomrule
\end{tabular}
}
\caption{Comparison of HyperVL with other models on internal benchmarks. The highest scores are marked in bold and the second highest are underlined.}
\label{tbl:internal}
\end{table*}
To assess practical utility in real-world deployment, we evaluate several models on our internal benchmarks covering four critical business scenarios. As shown in Table \ref{tbl:internal}, despite having the smallest parameter size (1.8B), HyperVL demonstrates superior capability in high-level semantic understanding and creative generation, significantly outperforming larger competitors.

Our model maintains strong competitiveness in fundamental visual perception and structured processing tasks. In Intent Recognition and Recommendation, HyperVL achieves an impressive 94.0, outperforming Qwen3-VL and InternVL3, proving its reliability in capturing latent user intent. While optimizing for high-level reasoning, it maintains a solid baseline in UI Understanding and Structure Parsing. Although slightly lower than other models, its performance effectively balances structural perception with advanced cognitive capabilities, ensuring robust support for downstream interactive applications. 

HyperVL demonstrates exceptional performance in tasks requiring multimodal alignment and creativity. In the challenging Image-Text Creation task, our model ranks first with a score of 49.8, establishing a substantial lead over Qwen3-VL and outperforming the strong baseline AndesVL. This highlights its exceptional proficiency in generating context-aware, high-quality content. Additonally, in Image Relevance Ranking, HyperVL ranks first, surpassing specialized models like SAIL-VL2, which confirms its precision in fine-grained visual-semantic matching essential for search and recommendation systems.

\subsubsection{Ablation Studies}

\begin{table*}[htbp]
\renewcommand{\arraystretch}{1.1}
\resizebox{\textwidth}{!}{%
\begin{tabular}{cccccccc}
\toprule
Methods & \textbf{AI2D\textsubscript{TEST}} & \textbf{ChartQA\textsubscript{TEST}} & \textbf{TextVQA\textsubscript{VAL}} & \textbf{DocVQA} & \textbf{OCRBench} & \textbf{Ch\_OCR} & \textbf{MMStar} \\
\midrule
HyperVL ViT$_{\text{L}}$  & 83.1 & 85.7 & 78.8 & 92.2 & 859 & 84.3 & 61.1 \\
\midrule
HyperVL                   & 76.8 & 81.2 & 75.5 & 90.8 & 808 & 79.8 & 58.1 \\
+ DCL                     & 81.8 & 83.8 & 75.7 & 91.3 & 830 & 80.4 & 59.2 \\
$\triangle$               & +5.0 & +2.6 & +0.2 & +0.5 & +22.0 & +0.6 & +1.1 \\
\bottomrule
\end{tabular}%
}
\caption{Impact dual consistency learning. +DCL means that HyperVL model is trained with dual consistency learning. Ch\_OCR refers to ChineseOCRBench.}
\label{tbl:dual}
\end{table*}

\textbf{Effectiveness  of Dual Consistency Learning.} To validate the efficacy of the proposed Dual Consistency Learning (DCL) strategy, we compare the DCL-enhanced model against the baseline HyperVL and the large-scale teacher variant (HyperVL ViT\textsubscript{L}). As detailed in Table \ref{tbl:dual}, the integration of DCL yields substantial performance gains across all benchmarks. Notably, the model achieves a significant improvement of +5.0 on AI2D and +2.6 on ChartQA, alongside a remarkable +22.0 point increase on OCRBench. These results strongly suggest that our strategy effectively facilitates the transfer of fine-grained visual knowledge, enabling the lightweight model to bridge the performance gap with larger vision encoders.

\textbf{Effectiveness of Visual Resolution Compressor.} 
We evaluate the Visual Resolution Compressor (VRC) on its ability to balance computational cost with model performance under strict on-device constraints. With a marginal overhead of only 2 ms, the VRC is virtually cost-free compared to standard vision encoders yet delivers significant throughput improvements. Experimental results in Table \ref{tbl:compressor_efficiency} confirm its efficacy: HyperVL achieves a 20.2\% reduction in visual tokens while maintaining 98.7\% of the baseline performance. The benefits are even more pronounced on Qwen3-VL-2B, which sees a 29.3\% token reduction with 98.4\% performance retention.

Furthermore, the VRC demonstrates content-adaptive capabilities. It intelligently adjusts compression levels based on image complexity: aggressively compressing high-redundancy inputs like DocVQA (49.5\%–63.4\% reduction), while preserving full resolution for detail-dense tasks like ChartQA (~2\%–3\% reduction) to ensure readability. For mixed scenarios like OCRBench, it adopts a balanced strategy (~30\%). In summary, the VRC provides a stable and efficient trade-off, accelerating inference for large inputs without compromising accuracy in detail-sensitive applications.

\begin{table*}[htbp]
\renewcommand{\arraystretch}{1.1} %
\resizebox{\textwidth}{!}{%
\begin{tabular}{lccccccc|c}
\toprule
\textbf{Model} & \textbf{AI2D\textsubscript{TEST}} & \textbf{ChartQA\textsubscript{TEST}} & \textbf{TextVQA\textsubscript{VAL}} & \textbf{DocVQA} & \textbf{OCRBench} & \textbf{Ch\_OCR} & \textbf{MMStar} & \textbf{Avg.} \\
\midrule
\multicolumn{9}{l}{\textbf{HyperVL}} \\
\midrule
Baseline & 81.8 & 83.8 & 75.7 & 91.3 & 830 & 80.4 & 59.2 & 100\% \\
+ VRC & 81.1 & 83.8 & 75.4 & 87.9 & 820 & 80.0 & 57.9 & 98.7\% \\
Compression Ratio & 14.9\% & 3.2\% & 4.0\% & 49.5\% & 30.9\% & 27.5\% & 11.3\% & 20.2\% \\
\midrule
\multicolumn{9}{l}{\textbf{HyperVL ViT\textsubscript{L}}} \\
\midrule
Baseline & 83.1 & 85.7 & 78.8 & 92.2 & 859 & 84.3 & 61.1 & 100\% \\
+ VRC & 83.1 & 85.5 & 78.7 & 88.9 & 846 & 82.7 & 61.7 & 99.1\% \\
Compression Ratio & 14.9\% & 3.2\% & 4.0\% & 49.5\% & 30.9\% & 27.5\% & 11.3\% & 20.2\% \\
\midrule
\multicolumn{9}{l}{\textbf{Qwen3-VL-2B}} \\
\midrule
Baseline & 76.9 & 77.6$^{*}$ & 79.8$^{*}$ & 93.3 & 858 & 78.0$^{*}$ & 58.3 & 100\% \\
+ VRC & 76.9 & 77.6 & 77.82 & 88.9 & 831 & 76.8 & 58.8 & 98.4\% \\
Compression Ratio & 14.4\% & 2.2\% & 36.0\% & 63.4\% & 35.0\% & 46.5\% & 7.9\% & 29.3\% \\
\bottomrule
\end{tabular}
}
\caption{Performance and token compression ratio of the visual resolution compressor across different models and benchmarks. Ch\_OCR refers to ChineseOCRBench.}
\label{tbl:compressor_efficiency}
\end{table*}

% \subsubsection{Impact of Reinforcement Learning}

\subsubsection{On-Device Evaluation}

\begin{figure*}[htp!]
\centering

\begin{subfigure}[b]{0.48\textwidth}
    \centering
    \includegraphics[width=\linewidth]{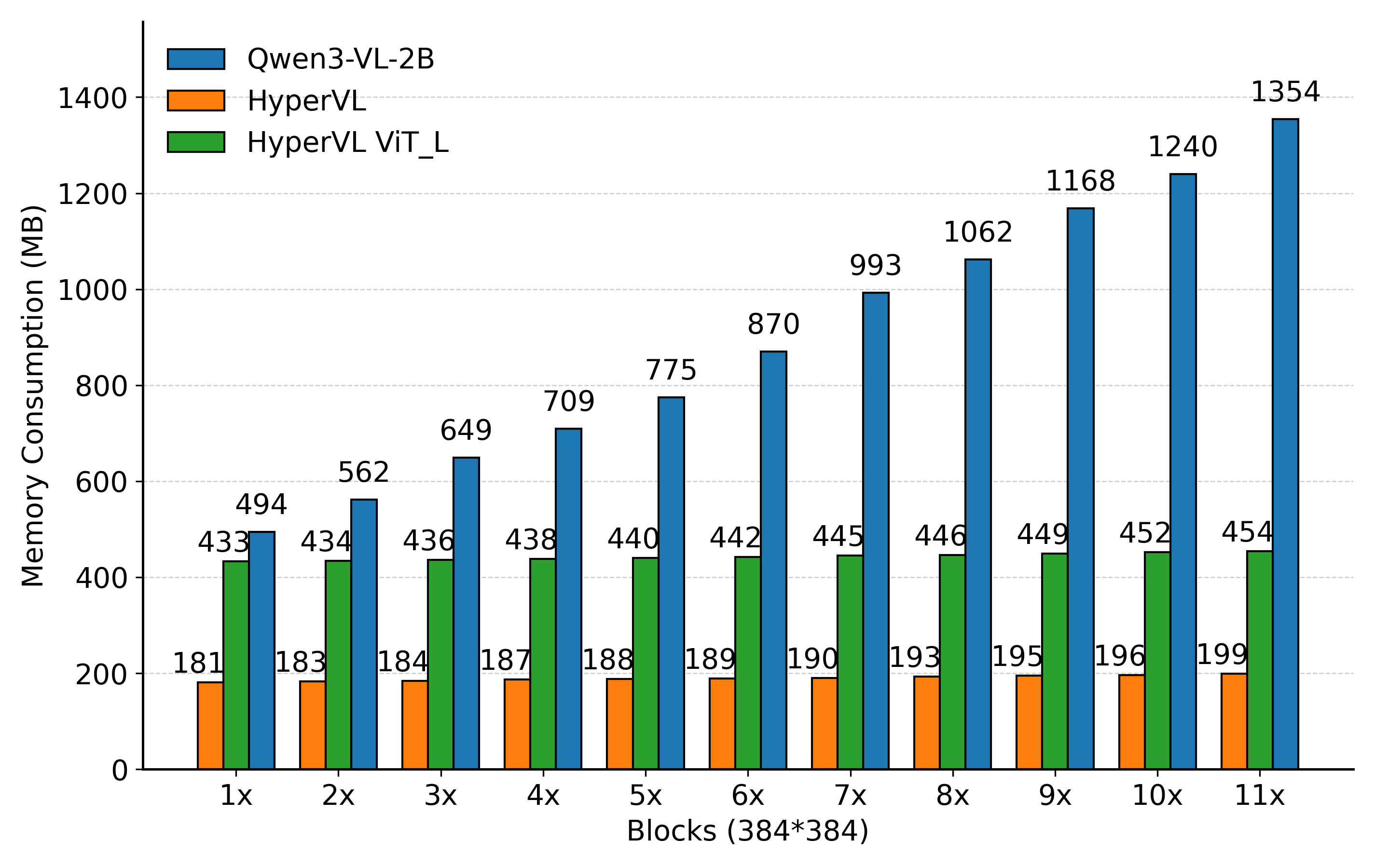}
    % \caption{Comparison of ViTs memory consumption (MB).}
    \label{fig:1}
\end{subfigure}
\hfill
\begin{subfigure}[b]{0.48\textwidth}
    \centering
    \includegraphics[width=\linewidth]{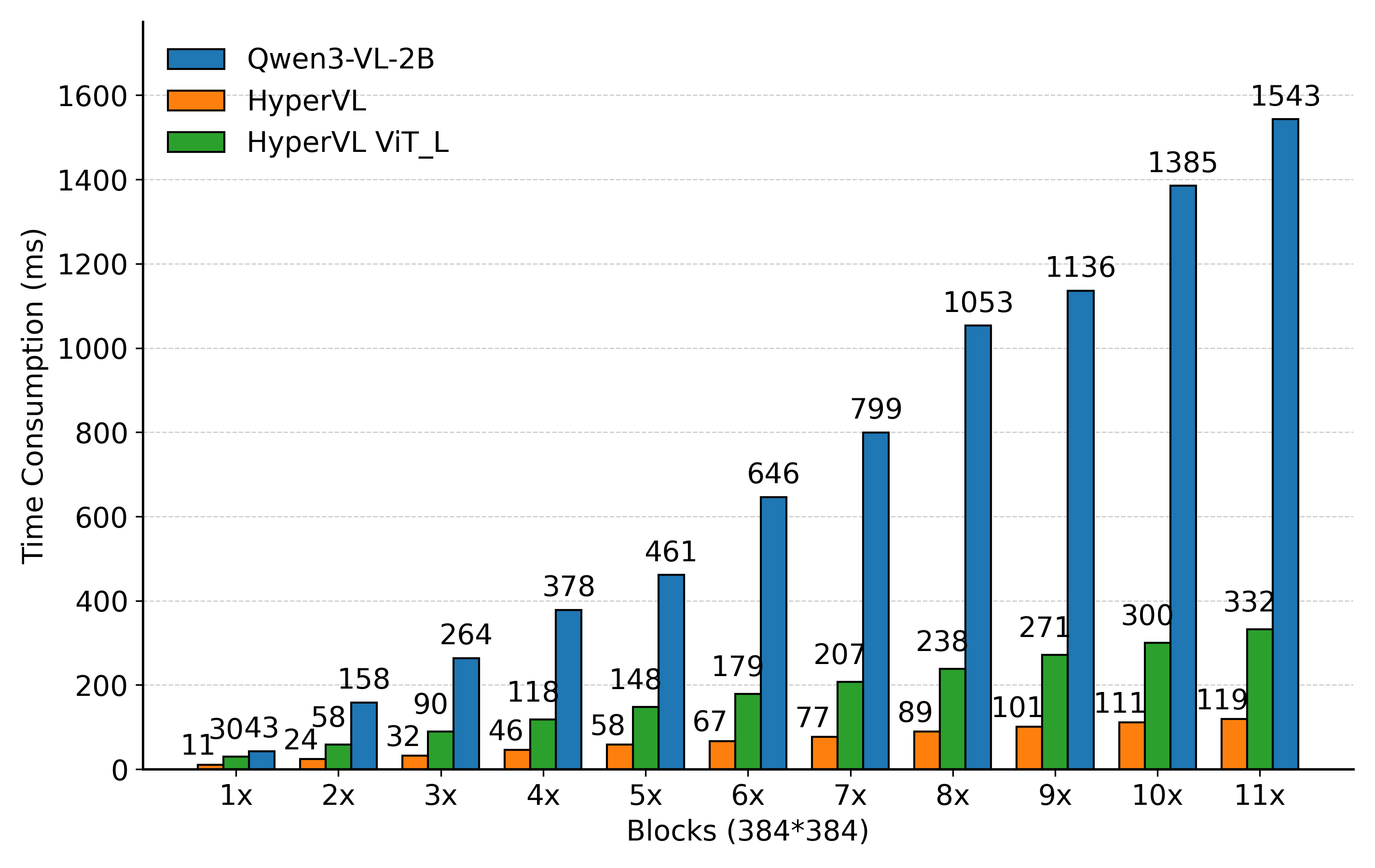}
    % \caption{Comparison of ViTs inference latency (ms).}
    \label{fig:2}
\end{subfigure}

\caption{Efficiency comparison of on-device ViTs cross different image resolutions on Qualcomm 8750 Platform. Left: memory consumption (MB) comparison of diffenet ViTs. Right: inference latency (ms) comparison of different ViTs.}
\label{fig:on-device comparsion}
\end{figure*}

\textbf{Efficiency Analysis.} 
To evaluate the feasibility of running MLLMs on resource-constrained on-device platforms, we conducted a comparative analysis of inference latency and memory overhead between HyperVL, HyperVL ViT\textsubscript{L}, and the baseline Qwen3-VL-2B. The results demonstrate the significant advantages of our architecture in terms of scalability and resource efficiency.

As illustrated in Figure \ref{fig:on-device comparsion}, HyperVL demonstrates superior inference efficiency and linear scalability compared to the baseline. While Qwen3-VL-2B suffers from a dramatic increase in latency as the image resolution increases. HyperVL maintains a highly efficient processing rate. This represents an approximate 12.9$\times$ speedup, ensuring real-time responsiveness even for high-resolution inputs. Crucially, regarding resource consumption, unlike the dynamic memory growth of Qwen3-VL (peaking at 1354 MB) HyperVL overall maintains a constant memory usage regardless of input complexity, achieving a 6.8$\times$ reduction in peak memory overhead. 

The superior efficiency of HyperVL stems from its alignment with the hardware characteristics of Qualcomm NPUs. Standard ViTs using QNN deployment require static allocation of intermediate buffers; as resolution increases, large attention matrices exceed the limited on-chip VTCM, triggering frequent, costly data swaps with external DDR and causing exponential latency growth. In contrast, our serial processing strategy fundamentally alters this scaling law. By processing small, fixed-size chunks, we ensure that intermediate activations consistently fit within the high-speed VTCM. This minimizes memory bandwidth bottlenecks and maintains a constant memory footprint, ensuring that inference latency scales linearly rather than exponentially with image resolution.

% These results confirm that our architecture effectively decouples visual resolution from computational cost, making it highly suitable for on-device deployment on resource-constrained platforms.

\begin{table*}[htbp]
\centering
\renewcommand{\arraystretch}{1.1}
\resizebox{\textwidth}{!}{%
\begin{tabular}{l|ccccccc}
\toprule
\textbf{Quantization} & \textbf{AI2D\textsubscript{TEST}} & \textbf{ChartQA\textsubscript{TEST}} & \textbf{TextVQA\textsubscript{VAL}} & \textbf{DocVQA} & \textbf{InfoVQA} & \textbf{OCRBench} & \textbf{Ch\_OCR} \\
\midrule
floating point         & 81.8 & 83.8    & 75.7         & 91.3        & 70.2    & 830      & 80.4    \\
w4a16        & 80.6 & 83.2    & 73.5         & 91.2        & 68.1    & 815      & 79.8   \\
\bottomrule
\end{tabular}
}
\caption{Performance comparison of quantization.}
\label{tbl:on-device accuracy}
\end{table*}

\textbf{Quantization Accuracy Analysis.} To validate the feasibility of on-device deployment, we evaluate the model's performance under W4A16 quantization (4-bit weights, 16-bit activations). As shown in Table \ref{tbl:on-device accuracy}, our model demonstrates exceptional robustness to low-bit precision.  Notably, on DocVQA, the quantized model retains virtually all of its FP16 performance (91.3 $\rightarrow$ 91.2), exhibiting a negligible drop of only 0.1 points. Similarly, in ChartQA, the performance remains highly stable (83.8 $\rightarrow$ 83.2). Even on aggregate benchmarks like OCRBench, the model maintains 98.2\% of its original capability (830 $\rightarrow$ 815). This result confirms that the proposed quantization scheme achieves a highly favorable trade-off, significantly reducing memory bandwidth requirements while preserving accuracy.

%% file: sec/5_conclusion.tex
\section{Conclusion}

We present HyperVL, an efficient multimodal large language model designed for on-device inference. By addressing the bottleneck of high-resolution visual encoders through image tiling, visual resolution compressor, and dual consistency learning, HyperVL achieves state-of-the-art performance on multiple multimodal benchmarks while significantly reducing latency and memory usage on mobile devices. The model demonstrates strong generalization for on-device tasks such as UI understanding and parsing, intent recommendation, and image-text creation.

Future work includes exploring adaptive token sparsity and attention pruning to further improve efficiency, extending the model to video and interactive scenarios, and incorporating user-adaptive personalization to enhance relevance and resource-aware performance in real-world mobile applications.

%% file: sec/7_authors.tex
\section{Contributor}
Yuchen Liu\textsuperscript{*}, Kaiyang Han\textsuperscript{*}, Zhiqiang Xia\textsuperscript{*}, Yuhang Dong\textsuperscript{*}, Chen Song\textsuperscript{*}, Kangyu Tang\textsuperscript{*}\footnotetext{* Equal contribution.}, Jiaming Xu$^\dagger$\footnotetext{$^\dagger$ Project leader, \textcolor{gray}\faEnvelope{ xujiaming1@xiaomi.com}}, Xiushi Feng, WenXuan Yu, Li Peng, Mingyang Wang, Kai Wang, Changpeng Yang, Yang Li, Haoyu Lu, Hao Wang, Bingna Xu, Guangyao Liu, Long Huang, Kaibin Guo, Jinyang Wu, Dan Wu, Hongzhen Wang, Peng Zhou, Shuai Nie, Shande Wang, Runyu Shi, Ying Huang 

%% file: sec/appendixes.tex
% \documentclass[a4paper,12pt]{article}

% % 宏包
% \usepackage{appendix}
% \usepackage{graphicx}
% \usepackage{array}
% \usepackage{booktabs}
% \usepackage{ragged2e}
% \usepackage{enumitem}
% \usepackage{caption}
% \usepackage{amsmath}
% \usepackage{CJKutf8}

% \begin{document}

% \begin{appendices}
\section{Qualitative Examples}
In this section, we present a collection of qualitative examples that illustrate the breadth of capabilities supported by \textbf{HyperVL}. These examples span a variety of multimodal reasoning and perception tasks, including mathematical computation, structured information extraction, geometric reasoning, interface-level planning, image description with character recognition, common-sense inference, and knowledge-based understanding.

Table~\ref{tab:math} provides an example of mathematical computation derived from visual input, demonstrating the model’s ability to perform systematic numerical reasoning based on images containing equations or quantitative information. Table~\ref{tab:chart} presents an example of chart content extraction, where the model effectively interprets structured visual representations such as plots and diagrams. Table~\ref{tab:geometry} illustrates an example of geometric reasoning, showing the model’s capacity to integrate diagrammatic cues with symbolic reasoning.

Table~\ref{tab:gui} offers an example of GUI understanding and planning, highlighting how the model interprets graphical user interface elements and generates coherent action plans. Table~\ref{tab:image-text} demonstrates an example of image description combined with character extraction, reflecting the model’s ability to simultaneously comprehend visual scenes and transcribe embedded text. 

Table~\ref{tab:commonsense} provides an example of common-sense reasoning grounded in real-world scenarios. Finally, Table~\ref{tab:knowledge} showcases an example of knowledge-based reasoning in the artistic domain, where HyperVL interprets an artwork and exhibits both its grasp of art-specific knowledge and its capacity for extended contextual associations.
\begin{table}[htbp]
\centering
\begin{tabular}{
  >{\centering\arraybackslash}m{0.35\textwidth}
  >{\raggedright\arraybackslash}m{0.65\textwidth}
}
\toprule
% 左列：图片垂直居中
\raisebox{-3\totalheight}{%
  \includegraphics[width=\linewidth]{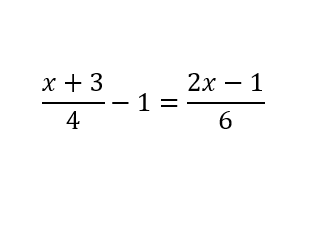}%
}
&
% 右列：minipage 顶部去空行
\begin{minipage}[t]{\linewidth}
% \vspace{-\baselineskip}% 轻微收掉顶部空白
\begin{CJK*}{UTF8}{gbsn}
\noindent
\textbf{Prompt:}\\ 请解决这个问题。\\[0.1em]
\textbf{Model Response:}\\[0.1em]
好的，我们来一步步解决这个方程。

**原方程：**
$$ \frac{x + 3}{4} - 1 = \frac{2x - 1}{6} $$

**步骤 1：消除分母**
为了消去分母，我们可以找到 4 和 6 的最小公倍数（LCM）。4 和 6 的最小公倍数是 12。
将方程两边同时乘以 12，以消除分母。

$$ 12 \times \left( \frac{x + 3}{4} - 1 \right) = 12 \times \frac{2x - 1}{6} $$

**步骤 2：简化方程**
计算左边：
$$ 12 \times \frac{x + 3}{4} - 12 \times 1 = 3(x + 3) - 12 $$
$$ 3x + 9 - 12 = 3x - 3 $$

计算右边：
$$ 12 \times \frac{2x - 1}{6} = 2(2x - 1) $$
$$ 4x - 2 $$

现在方程变为：
$$ 3x - 3 = 4x - 2 $$

**步骤 3：移项**
将含有 x 的项移到一边，常数项移到另一边。
从两边减去 3x：
$$ -3 = x - 2 $$

两边同时加上 2：
$$ -3 + 2 = x $$
$$ x = -1 $$

**步骤 4：验证解**
将 x = -1 代入原方程进行验证：
左边：$$ \frac{-1 + 3}{4} - 1 = \frac{2}{4} - 1 = \frac{1}{2} - 1 = -\frac{1}{2} $$
右边：$$ \frac{2(-1) - 1}{6} = \frac{-2 - 1}{6} = \frac{-3}{6} = -\frac{1}{2} $$
两边相等，验证正确。

**最终答案：**
$$ x = -1 $$
\end{CJK*}
\end{minipage}
\\
\bottomrule
\end{tabular}

\vspace{0.8em}
\caption{\begin{CJK*}{UTF8}{gbsn} Mathematical reasoning results.
 \end{CJK*}}
\label{tab:math}
\end{table}

\begin{table}[htbp]
\centering
\begin{tabular}{
  >{\centering\arraybackslash}m{0.35\textwidth}
  >{\raggedright\arraybackslash}m{0.65\textwidth}
}
\toprule
% 左列：图片垂直居中
\raisebox{-1.15\totalheight}{%
  \includegraphics[width=\linewidth]{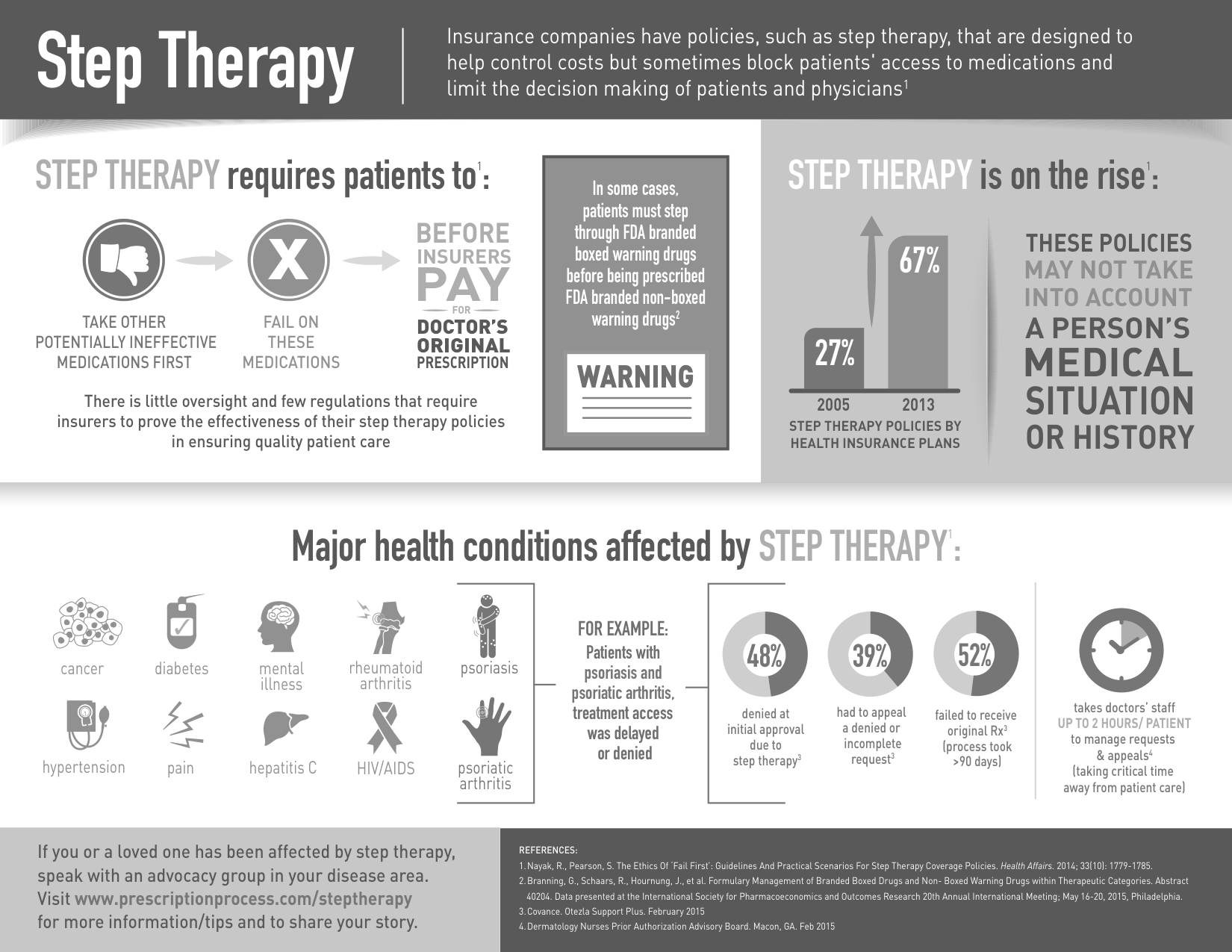}%
}
&
% 右列：minipage 顶部去空行
\begin{minipage}[t]{\linewidth}
% \vspace{-\baselineskip}% 轻微收掉顶部空白
\begin{CJK*}{UTF8}{gbsn}
\noindent
\textbf{Prompt:}\\ What's the increase of the step therpy between 2005 to 2013? Please calculate it step by step.\\[0.25em]
\textbf{Model Response:}\\[0.25em]
To calculate the increase of step therapy between 2005 and 2013, we need to subtract the percentage of step therapy in 2005 from the percentage in 2013.\\
Percentage of step therapy in 2005: 27\%\\
Percentage of step therapy in 2013: 67\%\\
Increase = (Percentage in 2013 - Percentage in 2005)\\
Increase = (67\% - 27\%)\\
Increase = 40\%\\
Therefore, the increase of step therapy between 2005 and 2013 is 40\%.
\end{CJK*}
\end{minipage}
\\
\bottomrule
\end{tabular}

\vspace{0.8em}
\caption{\begin{CJK*}{UTF8}{gbsn}  Visual chart interpretation and data extraction.
 \end{CJK*}}
\label{tab:chart}
\end{table}

\begin{table}[htbp]
\centering
\begin{tabular}{
  >{\centering\arraybackslash}m{0.35\textwidth}
  >{\raggedright\arraybackslash}m{0.65\textwidth}
}
\toprule
% 左列：图片垂直居中
\raisebox{-1\totalheight}{%
  \includegraphics[width=\linewidth]{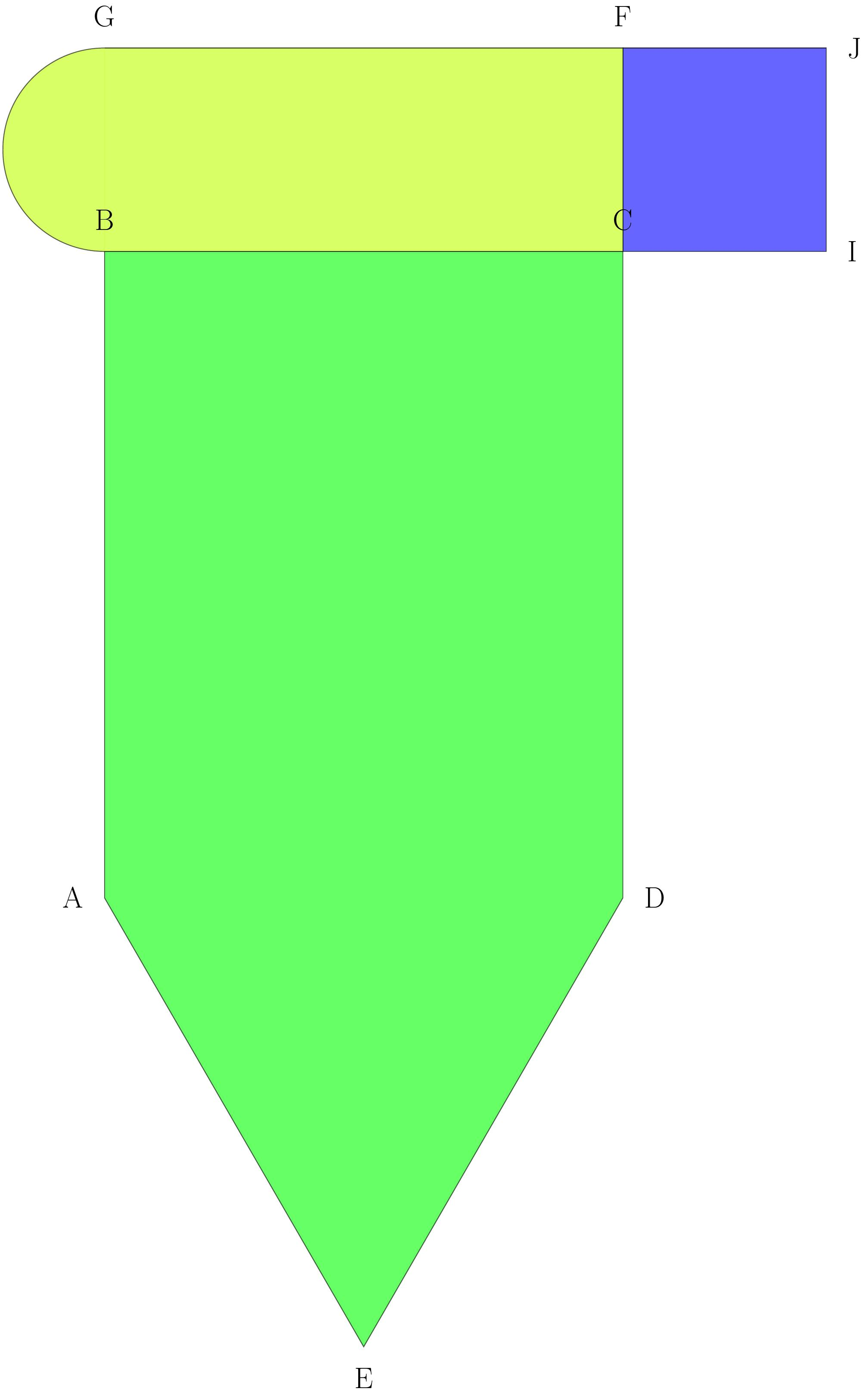}%
}
&
% 右列：minipage 顶部去空行
\begin{minipage}[t]{\linewidth}
% \vspace{-\baselineskip}% 轻微收掉顶部空白
\begin{CJK*}{UTF8}{gbsn}
\noindent
\textbf{Prompt:}\\ 
If the ABCDE shape is a combination of a rectangle and an equilateral triangle, the perimeter of the ABCDE shape is 84, the BCFG shape is a combination of a rectangle and a semi-circle, the perimeter of the BCFG shape is 46 and the area of the CIJF square is 36, compute the length of the AB side of the ABCDE shape. Assume $\pi=3.14$. Round computations to 2 decimal places.\\[0.25em]
\textbf{Model Response:}\\[0.25em]
The area of the CIJF square is 36, so the length of the CF side is $\sqrt{36} = 6$. The perimeter of the BCFG shape is 46 and the length of the CF side is 6, so $2 * OtherSide + 6 + \frac{6 * 3.14}{2} = 46$. So $2 * OtherSide = 46 - 6 - \frac{6 * 3.14}{2} = 46 - 6 - \frac{18.84}{2} = 46 - 6 - 9.42 = 30.58$. Therefore, the length of the BC side is $\frac{30.58}{2} = 15.29$. The side of the equilateral triangle in the ABCDE shape is equal to the side of the rectangle with length 15.29 so the shape has two rectangle sides with equal but unknown lengths, one rectangle side with length 15.29, and two triangle sides with length 15.29. The perimeter of the ABCDE shape is 84 so $2 * UnknownSide + 3 * 15.29 = 84$. So $2 * UnknownSide = 84 - 45.87 = 38.13$, and the length of the AB side is $\frac{38.13}{2} = 19.07$. Therefore the final answer is 19.07.

\end{CJK*}
\end{minipage}
\\
\bottomrule
\end{tabular}

\vspace{0.8em}
\caption{\begin{CJK*}{UTF8}{gbsn} An example of geometric reasoning. \end{CJK*}}
\label{tab:geometry}
\end{table}

\begin{table}[htbp]
\centering
\begin{tabular}{
  >{\centering\arraybackslash}m{0.35\textwidth}
  >{\raggedright\arraybackslash}m{0.65\textwidth}
}
\toprule
% 左列：图片垂直居中
\raisebox{-0.5\totalheight}{%
  \includegraphics[width=\linewidth]{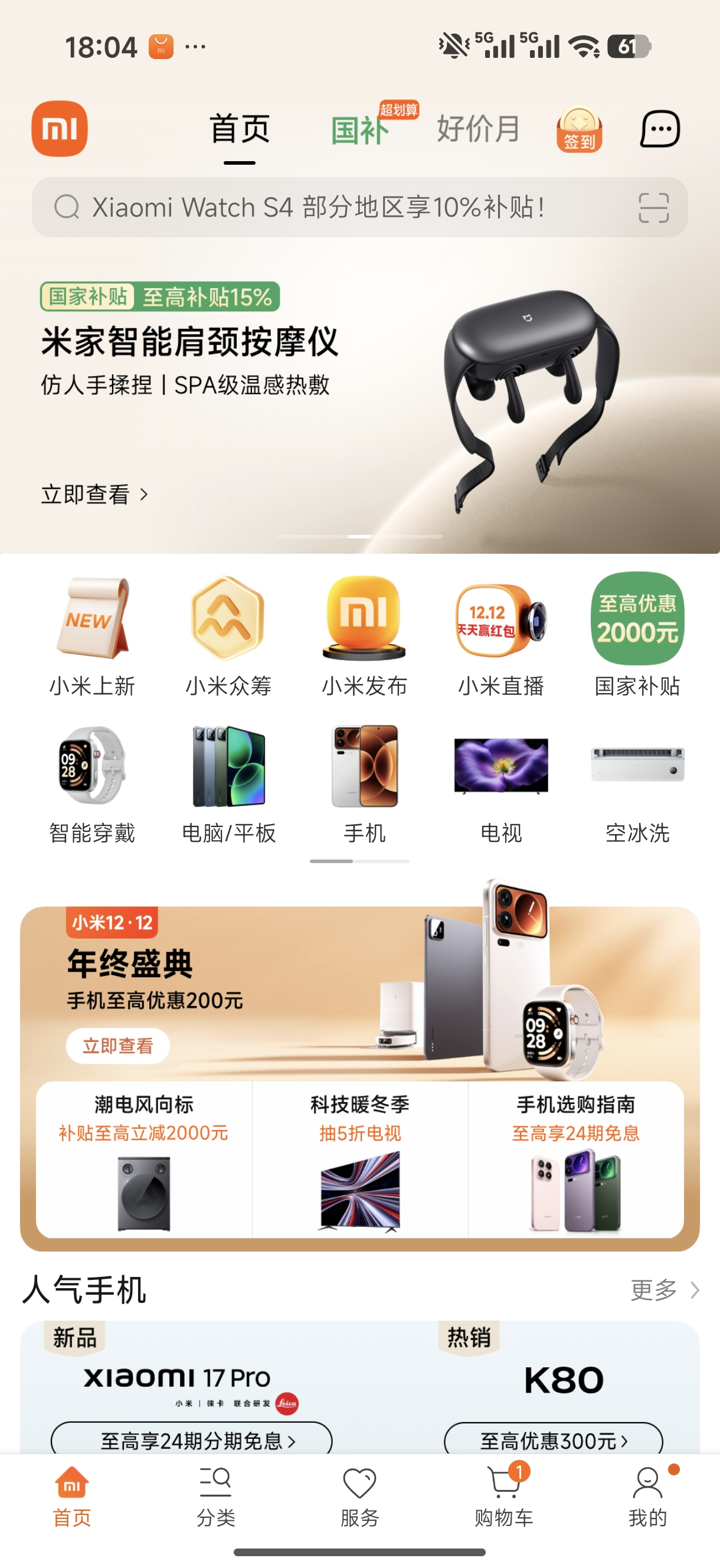}%
}
&
% 右列：minipage 顶部去空行
\begin{minipage}[t]{\linewidth}
% \vspace{-\baselineskip}% 轻微收掉顶部空白
\begin{CJK*}{UTF8}{gbsn}
\noindent
\textbf{Prompt:}\\ 
怎样购买手机最划算？\\[0.25em]
\textbf{Model Response:}\\[0.25em]
根据图片中的信息，购买手机可以享受“手机至高优惠200元”的活动，具体优惠金额和购买渠道可以在“年终盛典”部分查看。此外，还可以通过“手机选购指南”获取更多购买建议和优惠信息。

\end{CJK*}
\end{minipage}
\\
\bottomrule
\end{tabular}

\vspace{0.8em}
\caption{\begin{CJK*}{UTF8}{gbsn} GUI understanding and task planning performed.
\end{CJK*}}
\label{tab:gui}
\end{table}

\begin{table}[htbp]
\centering
\begin{tabular}{
  >{\centering\arraybackslash}m{0.35\textwidth}
  >{\raggedright\arraybackslash}m{0.65\textwidth}
}
\toprule
% 左列：图片垂直居中
\raisebox{-1.3\totalheight}{%
  \includegraphics[width=\linewidth]{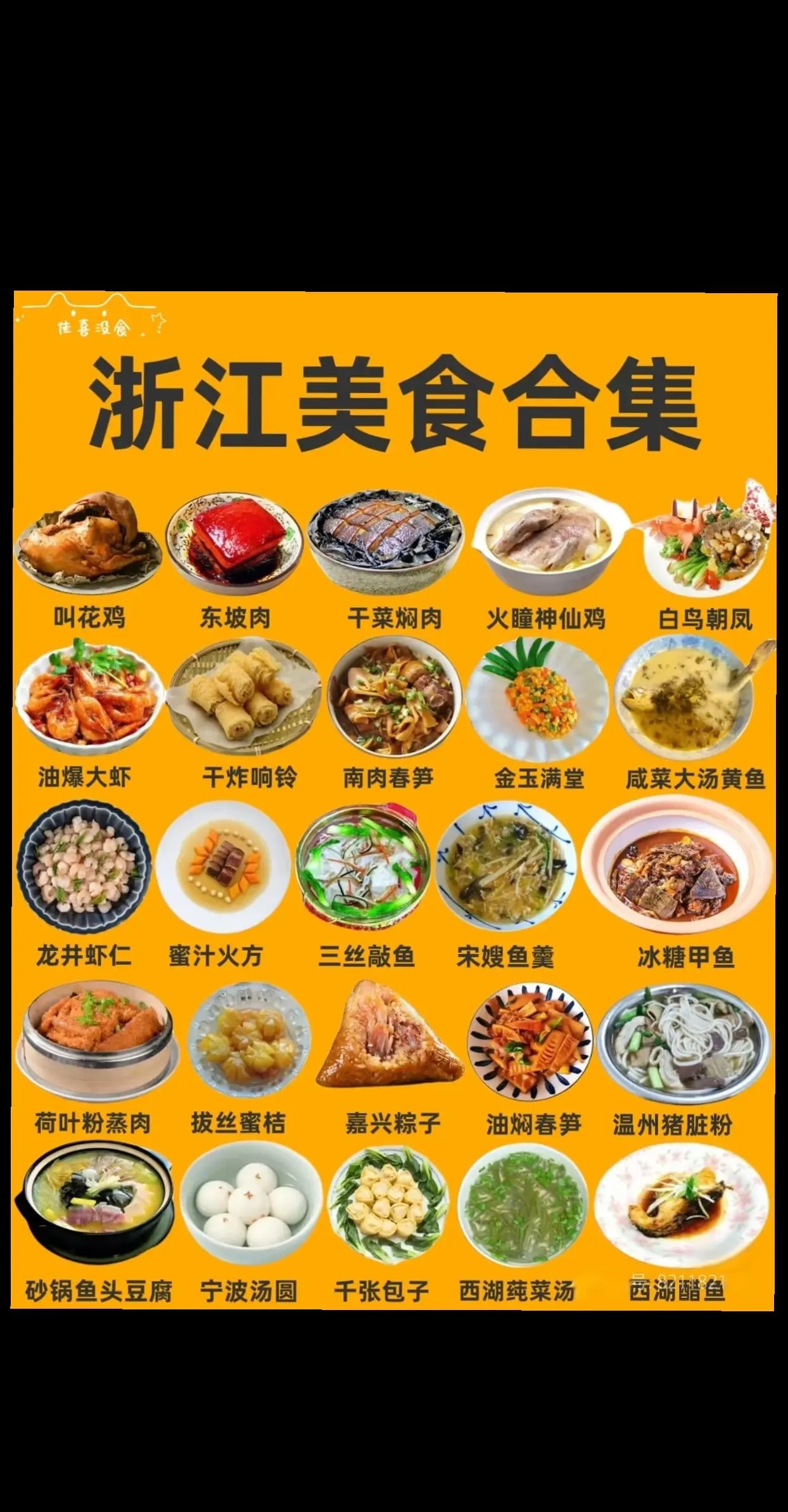}%
}
&
% 右列：minipage 顶部去空行
\begin{minipage}[t]{\linewidth}
% \vspace{-\baselineskip}% 轻微收掉顶部空白
\begin{CJK*}{UTF8}{gbsn}
\noindent
\textbf{Prompt:}\\ 
详细描述一下这张图片\\[0.25em]
\textbf{Model Response:}\\[0.25em]
这张图片展示了一张名为“浙江美食合集”的海报，背景为黄色。海报上展示了多种浙江特色菜肴的图片和名称，这些菜肴被排列成五行，每行有五道菜。\\\\第一行从左到右分别是：\\- 叫花鸡\\- 东坡肉\\- 干菜焖肉\\- 火瞳神仙鸡\\- 白鸟朝凤\\\\第二行从左到右分别是：\\- 油爆大虾\\- 干炸响铃\\- 南肉春笋\\- 金玉满堂\\- 咸菜大汤黄鱼\\\\第三行从左到右分别是：\\- 龙井虾仁\\- 蜜汁火方\\- 三丝敲鱼\\- 宋嫂鱼羹\\- 冰糖甲鱼\\\\第四行从左到右分别是：\\- 荷叶粉蒸肉\\- 拔丝蜜桔\\- 嘉兴粽子\\- 油焖春笋\\- 温州猪脏粉\\\\第五行从左到右分别是：\\- 砂锅鱼头豆腐\\- 宁波汤圆\\- 千张包子\\- 西湖莼菜汤\\- 西湖醋鱼\\\\每道菜的图片都清晰地展示了菜肴的外观，旁边配有相应的中文名称。整体设计简洁明了，突出了浙江美食的多样性和特色。
\end{CJK*}
\end{minipage}
\\
\bottomrule
\end{tabular}

\vspace{0.8em}
\caption{\begin{CJK*}{UTF8}{gbsn} Image description and text extraction.
 \end{CJK*}}
\label{tab:image-text}
\end{table}

\begin{table}[htbp]
\centering
\begin{tabular}{
  >{\centering\arraybackslash}m{0.35\textwidth}
  >{\raggedright\arraybackslash}m{0.65\textwidth}
}
\toprule
% 左列：图片垂直居中
\raisebox{-0.6\totalheight}{%
  \includegraphics[width=\linewidth]{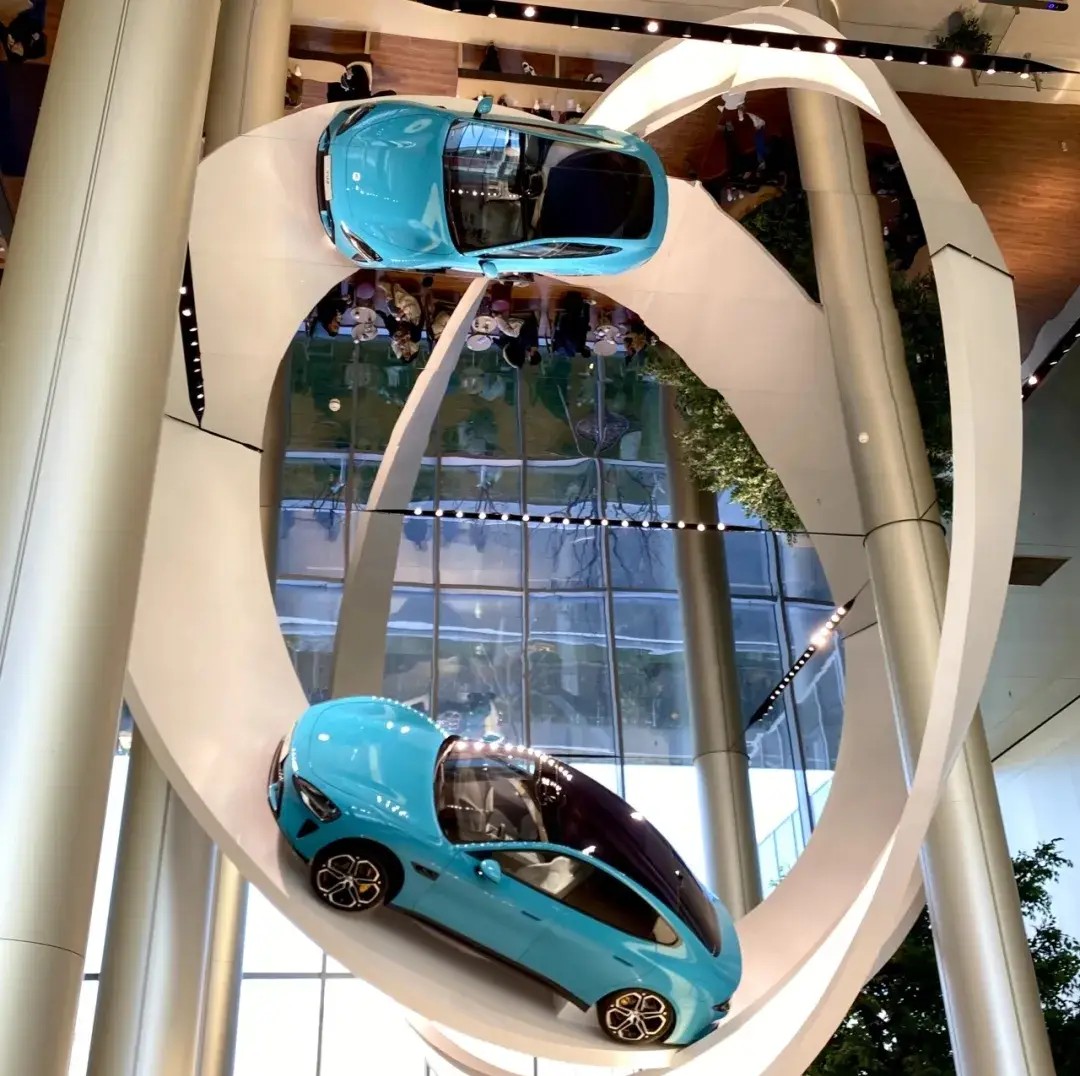}%
}
&
% 右列：minipage 顶部去空行
\begin{minipage}[t]{\linewidth}
% \vspace{-\baselineskip}% 轻微收掉顶部空白
\begin{CJK*}{UTF8}{gbsn}
\noindent
\textbf{Prompt:}\\ 
图中实际有几辆汽车\\[0.25em]
\textbf{Model Response:}\\[0.25em]
图中实际上只有一辆汽车。这辆车被设计成悬浮在空中，看起来像是有两辆车，但实际上只有一辆车。

\end{CJK*}
\end{minipage}
\\
\bottomrule
\end{tabular}

\vspace{0.8em}
\caption{\begin{CJK*}{UTF8}{gbsn} Common‑sense reasoning on real‑world scenarios.
 \end{CJK*}}
\label{tab:commonsense}
\end{table}

\begin{table}[htbp]
\centering
\begin{tabular}{
  >{\centering\arraybackslash}m{0.35\textwidth}
  >{\raggedright\arraybackslash}m{0.65\textwidth}
}
\toprule
% 左列：图片垂直居中
\raisebox{-0.9\totalheight}{%
  \includegraphics[width=\linewidth]{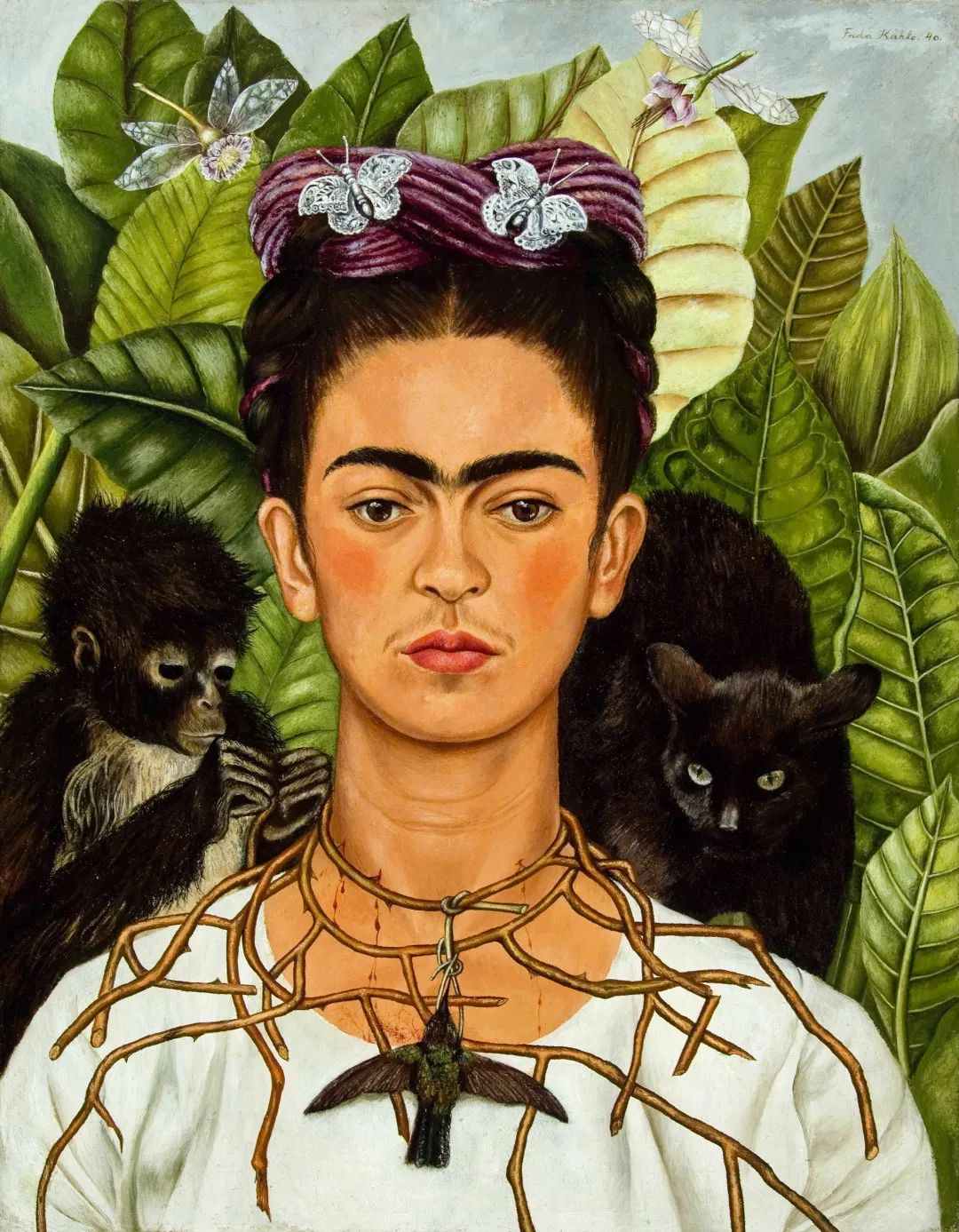}%
}
&
% 右列：minipage 顶部去空行
\begin{minipage}[t]{\linewidth}
% \vspace{-\baselineskip}% 轻微收掉顶部空白
\noindent
\textbf{Prompt:}\\ 
Please introduce the author of this painting\\[0.25em]
\textbf{Model Response:}\\[0.25em]
The painting you've shown is a modern interpretation of Frida Kahlo's self-portraits, specifically drawing inspiration from her iconic style and themes. Frida Kahlo was a Mexican artist known for her self-portraits, which often depicted her physical and emotional pain, as well as her cultural heritage. Her works frequently included vibrant colors, symbolic imagery, and a blend of realism and surrealism.\\
In this particular piece, the artist has incorporated elements such as butterflies, a monkey, a cat, and a hummingbird, which are common motifs in Kahlo's work. The background features lush green leaves, adding to the organic and natural feel of the painting.\\
\end{minipage}
\\
\bottomrule
\end{tabular}
\vspace{0.8em}
\caption{Knowledge‑based visual reasoning.}
\label{tab:knowledge}
\end{table}

% \end{appendices}
% \end{document}